%% file: main.tex
\documentclass[10pt,twocolumn,letterpaper]{article}

\usepackage[pagenumbers]{cvpr}%

\makeatletter
\@namedef{ver@everyshi.sty}{}
\makeatother
\usepackage{tikz}

\usepackage{graphicx}
\usepackage{amsmath}
\usepackage{amssymb}
\usepackage{booktabs}

\usepackage{times}
\usepackage{epsfig} 
\usepackage{moresize}
\usepackage{amsfonts} 
\usepackage{bbm} 
\usepackage{mathtools} 
\usepackage{algorithm,algorithmic}
\usepackage{dsfont}
\usepackage{multirow} 
\usepackage{caption}  
\usepackage{siunitx}
\usepackage{pifont}%
\usepackage{makecell}
\usepackage{placeins} 
\usepackage{colortbl}
\usepackage{rotating}
\usepackage{array} 
\usepackage{xcolor}
\usepackage{verbatim}
\usepackage{units}
\usepackage{textcomp}
\usepackage{bbding}
\usepackage{mwe} 
\usepackage{wrapfig}
\usepackage{subcaption}
\usepackage{listings}
\usepackage{xr}
\PassOptionsToPackage{normalem}{ulem}
\usepackage{ulem}
\usepackage{float} 
\usepackage{graphics}
\usepackage[algo2e,ruled,lined,linesnumbered,commentsnumbered,longend]{algorithm2e}

\SetCommentSty{mycommfont}

\definecolor{darkgreen}{rgb}{0, 0.4, 0}
\definecolor{darkred}{rgb}{0.7, 0, 0}
\definecolor{darkblue}{rgb}{0.0, 0.0, 0.7}

\definecolor{emph_number}{rgb}{0.6, 0.05, 0}

\newcommand{\emphtable}[1]{\textcolor{emph_number}{#1}} 
\newcommand{\blue}[1]{\textcolor{darkblue}{#1}} 

\usepackage{stackengine,scalerel}

\newcommand\overstar[1]{\ThisStyle{\ensurestackMath{%
  \setbox0=\hbox{$\SavedStyle#1$}%
  \stackengine{0pt}{\copy0}{\kern.2\ht0\smash{\SavedStyle*}}{O}{c}{F}{T}{S}}}}

\newcommand{\myparagraph}[1]{\vspace{4pt}\noindent{\bf #1}}

\usepackage[hang]{footmisc}
\usepackage[pagebackref,breaklinks,colorlinks]{hyperref}

\usepackage[capitalize]{cleveref}
\crefname{section}{Sec.}{Secs.}
\Crefname{section}{Section}{Sections}
\Crefname{table}{Table}{Tables}
\crefname{table}{Tab.}{Tabs.}

\begin{document}

\title{\vspace{-2em}Discovering Class-Specific GAN Controls for Semantic Image Synthesis}

\author{Edgar Sch{\"o}nfeld$^1$%
\and
Julio Borges$^1$%
\and
Vadim Sushko$^1$%
\and
Bernt Schiele$^2$%
\and
Anna Khoreva$^{1,3}$ \vspace{0.4em}%
\and
$^1$Bosch Center for AI
\and
$^2$MPI for Informatics
\and
$^3$University of T{\"u}bingen
}

\maketitle

\begin{abstract}
  Prior work has extensively studied the latent space structure of GANs for unconditional image synthesis, enabling global editing of generated images by the unsupervised discovery of interpretable latent directions.
  However, the discovery of latent directions for conditional GANs for semantic image synthesis (SIS) has remained unexplored.  
  In this work, we specifically focus on addressing this gap. 
  We propose a novel optimization method for finding spatially disentangled class-specific directions in the latent space of pretrained SIS models. 
  We show that the latent directions found by our method can effectively control the local appearance of semantic classes, e.g., changing their internal structure, texture or color independently from each other. 
  Visual inspection and quantitative evaluation of the discovered GAN controls on various datasets demonstrate that our method discovers a diverse set of unique and semantically meaningful latent directions for class-specific edits.
  \end{abstract}

\input{tex/introduction.tex}

\input{tex/related_work.tex}

\input{tex/method.tex}

\input{tex/experiments.tex}

\input{tex/conclusion.tex}

{\small
\bibliographystyle{ieee_fullname}
\bibliography{egbib}
}

    \clearpage
    \appendix
    \input{appendix.tex}

\end{document}

%% file: tex/introduction.tex
\section{Introduction} \label{sec:introduction}
Semantic image synthesis (SIS) transforms user-specified semantic layouts to realistic images. 
Its applications range widely from image editing and content creation to synthetic data augmentation, where training data is generated to fulfill specific semantic requirements.
For SIS, GANs~\cite{goodfellow2014generative} have 
demonstrated their superiority in terms of the visual quality of synthesised images and their alignment to input semantic label maps~\cite{park2019semantic,schonfeld2020you,tan2021diverse,wang2021image,li2021collaging}.
Although some of GAN-based SIS models allow local appearance editing of single classes or regions in an image -- either by style transfer from a reference image~\cite{zhu2020sean,lee2020maskgan,tan2021diverse} or by sampling noise independently for specific image regions~\cite{schonfeld2020you,zhu2020semantically}, they are lacking the technique of enabling interpretable semantic changes for the specific class without reference image and user-in-the-loop supervision.   

\input{fig/teaser}

On the other hand, prior work has extensively studied the latent space of unconditional GANs~\cite{Goetschalckx2019GANalyzeTV,plumerault2019controlling,harkonen2020ganspace,shen2021closed,tzelepis2021warpedganspace,yuksel2021latentclr}, finding interpretable latent directions which activate distinctive factors of variations in the generation process in an unsupervised fashion, without exploiting reference images. 
Moving latent code(s) along a certain direction can result in domain-agnostic transformations, e.g. rotation or zooming \cite{voynov2020unsupervised,jahanian2019steerability,plumerault2019controlling}, or domain-specific alterations, e.g. age or nose length of a person \cite{Cherepkov2021NavigatingTG,wu2021stylespace,ling2021editgan,shen2020interpreting,Collins2020EditingIS}.
Despite their progress, it remains a challenge to find interpretable latent directions to control interactively the synthesis of specific semantic classes in the image without changing other image regions.
Since the above methods were designed specifically for unconditional GANs, they are not well suited 
to discover class-specific latent directions in the presence of semantic label maps, inherently given for SIS.

In this work, we address this limitation and study the latent space of conditional GANs designed for SIS, which to the best of our knowledge has not been explored previously.
In particular, making use of the label maps we devise a method to discover meaningful latent directions that only change a specific semantic class in the image. These directions can, for example, encode different designs of the facade for the building class or surfaces for the street class (Fig.~\ref{fig:teaser}), enabling the user to perform local semantic edits independently from the rest of the image.
Note that in recent state-of-the-art SIS GANs, the generator is already designed to be sensitive to spatial information~\cite{schonfeld2020you,zhu2020semantically}, allowing region-based noise sampling of specific classes. 

On this basis, we introduce a simple, efficient optimisation method to discover class-specific controls in pretrained SIS GANs, which we call \emph{Ctrl-SIS} (see Fig.~\ref{fig:method}). Our optimization objective is designed to ensure that the learnt latent directions are 1) diverse and different from each other (\emph{diversity loss}); 2) only affect the image area of the selected class, preserving the appearance of other areas (\emph{disentanglement loss}); and 3) induce the same semantic edits consistently across different initial latent codes and label maps containing the class (\emph{consistency loss}). See Sec.~\ref{sec:method} for more details. 
We demonstrate that GAN controls discovered automatically by Ctrl-SIS can effectively manipulate the appearance of the selected semantic class in specific ways, without affecting other classes in the image.
For example, we can change the house facade~(see Fig.~\ref{fig:teaser}), remove leaves from trees or cover mountains in snow (see Fig.~\ref{fig:tree_removal}).
Moreover, we can edit different classes jointly, e.g., alter both the building and the road in the street scene (see Fig.~\ref{fig:teaser}). 
Since we use only train-time optimization, instead of exhaustive search as in~\cite{wu2021stylespace} or test-time optimization as in~\cite{ling2021editgan,pajouheshgar2021optimizing,Zhu2021LowRankSI,zhu2022region}, our training time stays relatively fast compared to the former, while also allowing interactive image editing compared to the latter.
 
The evaluation of GAN control discovery methods is commonly left to subjective visual inspection. To address this, we introduce new metrics to quantitatively assess diversity, spatial disentanglement and consistency properties of learnt latent directions (see Sec.~\ref{subsec:exp_eval}). 
We compare Ctrl-SIS with other GAN control methods on different SIS GANs~\cite{schonfeld2020you,park2019semantic,wang2021image} on two datasets~\cite{zhou2017scene,caesar2018coco}. Our experiments show that latent directions found by prior 
methods adapted to SIS~\cite{harkonen2020ganspace,shen2021closed} lead to weaker class edits, comparable to random directions (see Sec.~\ref{subsec:exp_res}). In contrast, Ctrl-SIS finds directions that enable diverse and semantically meaningful class edits while maintaining high image quality. 

In summary, our contributions are as follows:
1) We propose Ctrl-SIS -- a method to discover interpretable latent controls for individual semantic classes in pretrained SIS GANs. To the best of our knowledge, the discovery of class-specific latent direction has not yet been addressed in the SIS literature. 2) We define diversity, consistency, and spatial disentanglement as desirable properties of class-specific latent controls and propose new metrics to quantify them.

%% file: fig/teaser.tex
\begin{figure}[htb]
\begin{centering}
\vspace{-1em}
\setlength{\tabcolsep}{0.1em}
\renewcommand{\arraystretch}{0}
\par\end{centering} 
\begin{centering} 
\hfill{}%
\begin{tabular}{c@{\hskip 0.05in}c@{\hskip 0.05in}c}

\multicolumn{3}{c}{
    \begin{tabular}{cc}
    \includegraphics[width=0.13\textwidth, height=0.075\textheight]{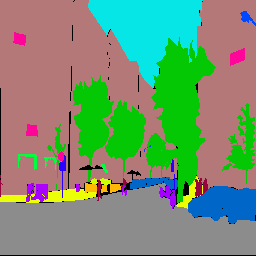} &
    \includegraphics[width=0.13\textwidth, height=0.075\textheight]{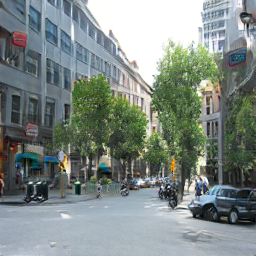} \tabularnewline 
    Label map  & Generated image 
    \end{tabular}
    
} \tabularnewline

\includegraphics[width=0.13\textwidth, height=0.075\textheight]{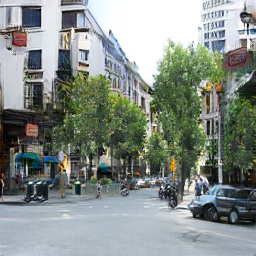} & 
\includegraphics[width=0.13\textwidth, height=0.075\textheight]{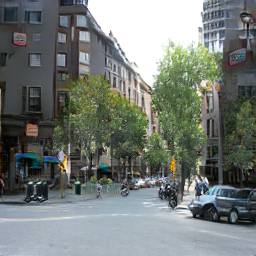} &
\includegraphics[width=0.13\textwidth, height=0.075\textheight]{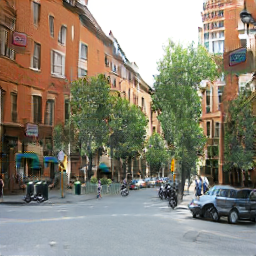} \tabularnewline 

 \multicolumn{3}{c}{Editing of class \textit{building}} \tabularnewline

 \includegraphics[width=0.13\textwidth, height=0.075\textheight]{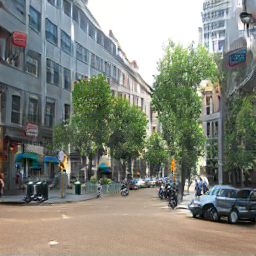}  &
 \includegraphics[width=0.13\textwidth, height=0.075\textheight]{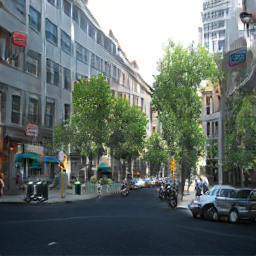} &  
 \includegraphics[width=0.13\textwidth, height=0.075\textheight]{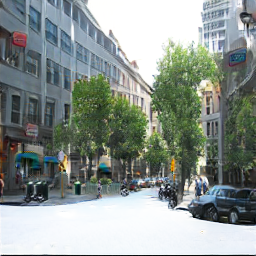} \tabularnewline 

  \multicolumn{3}{c}{Editing of class \textit{street}} \tabularnewline
  
 \includegraphics[width=0.13\textwidth, height=0.075\textheight]{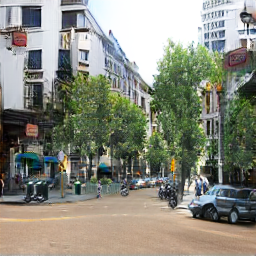} &
 \includegraphics[width=0.13\textwidth, height=0.075\textheight]{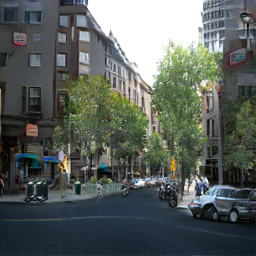} &  
 \includegraphics[width=0.13\textwidth, height=0.075\textheight]{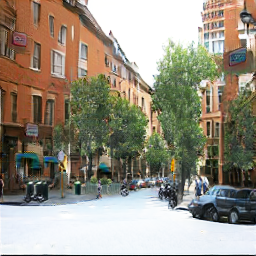} \tabularnewline  

  \multicolumn{3}{c}{Editing of class \textit{building} and \textit{street} jointly}
\end{tabular}
\hfill{}
\par\end{centering} 

\vspace{-0.5em}
\caption{Our Ctrl-SIS method learns class-specific directions in the latent space of a SIS model, which can be applied jointly for different semantic classes for local editing of the generated image.}
\label{fig:teaser} 
\vspace{-1em}
\end{figure}

%% file: tex/related_work.tex
\section{Related work} \label{sec:related_work}

\myparagraph{GAN models for SIS.}
\label{subsec:related_work_SIS} 
SIS GANs attracted a lot of attention for their application in controllable image synthesis \cite{park2019gaugan} and editing \cite{ntavelis2020sesame,tan2021diverse}. 
To achieve photorealism, Pix2Pix~\cite{isola2017image} used an encoder-decoder generator and a patch-based discriminator, providing label maps as input to the first layers of both networks.
SPADE \cite{park2019semantic} demonstrated that using label maps as input only to the first layer tends to weaken semantic conditioning and proposed to modulate intermediate generator layers via a spatially-adaptive normalization.
Follow-up works improved image quality through different ways of using label maps in the generator,
e.g., via other conditional normalizations \cite{tan2020semantic,ntavelis2020sesame,zhu2020sean,wang2021image}, conditional convolutions \cite{liu2019learning}, a label map encoder \cite{li2021collaging}, or by learning class-specific sub-generators~\cite{tang2020local}. 

Many SIS models struggled to achieve diversity, as the generator tended to ignore the input latent code~\cite{isola2017image,wang2018high}. To address this issue, SPADE \cite{park2019semantic} and SC-GAN \cite{wang2021image} utilized an image encoder to extract a global style vector from a reference image. 
By changing a reference image, these models can generate different images from the same label map. 
\cite{zhu2020sean, lee2020maskgan} further enabled class-wise style
transfers from the reference image.
\cite{zhu2020semantically} allowed local editing by controlling the appearance of different classes via separate latent codes using group convolutions. OASIS~\cite{schonfeld2020you} improved image diversity by feeding a 3D latent code tensor jointly with the label map into the conditional batch normalization layers, thus, enabling both global image editing as well as local region-based editing. 
For this reason, in this work we discover class-specific directions in the 3D latent spaces of already pre-trained generators of the state-of-the-art SIS models~\cite{park2019semantic,wang2021image,schonfeld2020you}. 
Since latent direction discovery is not applicable to the exemplar-based approaches, such as~\cite{zhu2020sean,lee2020maskgan}, we focus on non-exemplar-based models, that rely only on the input latent code to generate diverse images.

\input{fig/model.tex}
\myparagraph{GAN control discovery.} \label{subsec:related_work_latent}
It has been shown that the latent space of GANs frequently exhibits semantically relevant vector space arithmetic \cite{bau2019gandissect,Goetschalckx2019GANalyzeTV,jahanian2019steerability,voynov2020unsupervised,schwettmann2021toward}. 
However, finding steerable directions in the latent space is challenging due to its high dimensionality and the large variety of image semantics. 
Consequently, some works use human supervision \cite{schwettmann2021toward}, attribute predictors \cite{wu2021stylespace,shen2020interpreting} or predetermined visual transformations such as zooming and rotation \cite{jahanian2019steerability,plumerault2019controlling} to identify interpretable latent directions. 
As the dependence on supervision limits the practical use of these methods,
\cite{spingarn2020gan,voynov2020unsupervised,tzelepis2021warpedganspace,shen2021closed,harkonen2020ganspace,yuksel2021latentclr} investigated unsupervised discovery of GAN controls. 
GANSpace~\cite{harkonen2020ganspace} performed PCA on the intermediate generator features, discovering useful directions in the latent space resulting from layerwise perturbations along the principal directions. 
SeFa~\cite{shen2021closed} identified semantically meaningful directions in closed-form, through eigendecomposition of the generator weights. 
In contrast, \cite{yuksel2021latentclr,tzelepis2021warpedganspace} relied on gradient-based optimization. 
WarpedGanSpace~\cite{tzelepis2021warpedganspace} used an image classifier to discriminate among a fixed set of directions, while LatentCLR~\cite{yuksel2021latentclr} employed a contrastive loss optimizing directions to have orthogonal effects on the generator features. 
A common limitation of unsupervised methods is that the obtained latent directions are left to subjective visual inspection and manual identification of significant controls.

While the above work focused on finding latent directions for global image manipulation of unconditional GANs, 
our method finds class-specific latent directions for conditional SIS models. 
We pick state-of-the-art GANSpace and SeFa for comparison due to their simplicity, easy adaptation to SIS GAN models, and code availability.

\myparagraph{Local editing with GANs.} \label{subsec:local_edit}
Recent work enabled local image editing through optimization in the latent space on specific image regions~\cite{wu2021stylespace,Suzuki2018SpatiallyCI,ling2021editgan,pajouheshgar2021optimizing,zhu2022region,Zhu2021LowRankSI}.
EditGAN~\cite{ling2021editgan} modelled images and label maps jointly, requiring the user to modify the label map in order to perform the edit via latent space optimization. 
Image2StyleGAN++~\cite{abdal2020image2stylegan++} changed images locally via a masked style transfer or by re-encoding images edited with provided scribbles. 
LELSD \cite{pajouheshgar2021optimizing} proposed an area loss that, given a binary mask, optimizes changes only within the specified area. Yet, the found directions are still applied globally. 
ReSeFa~\cite{zhu2022region} proposed to optimize the change of pixel values with respect to the latent code, identifying latent variations in a user-specified image region.
The main limitation of the above work is that it requires test-time optimization, preventing the user from interactive image editing. 
In contrast, Ctrl-SIS is optimized end-to-end once to provide latent directions for interactive editing in the spirit of GANSpace and SeFa. It enables the user to perform image editing interactively, without the need for further supervision or mask area definitions. 

Ctrl-SIS differs from the aforementioned latent space optimization methods in two ways. First, Ctrl-SIS is specifically designed for SIS, making use of semantic label maps inherent in this task. Second, the discovered latent directions are class-specific. Several SIS models allow class-specific noise sampling~\cite{li2021collaging,zhu2020semantically,tan2021diverse,schonfeld2020you}, but do not provide a method to discover interpretable directions in their latent space. 
Lastly, existing methods for class-specific editing in SIS models manipulate classes using predetermined concepts~\cite{zhu2020sean,lee2020maskgan}, requiring a source image to extract a style. In contrast, our unsupervised approach allows users to browse through a set of discovered semantic concepts that the pretrained SIS GAN has learned.

%% file: fig/model.tex
\begin{figure*}[t]
\begin{centering}
\setlength{\tabcolsep}{0.1em}
\renewcommand{\arraystretch}{0}
\par\end{centering}
\begin{centering}
\vspace{-0.5em}
\includegraphics[width=0.85\textwidth]{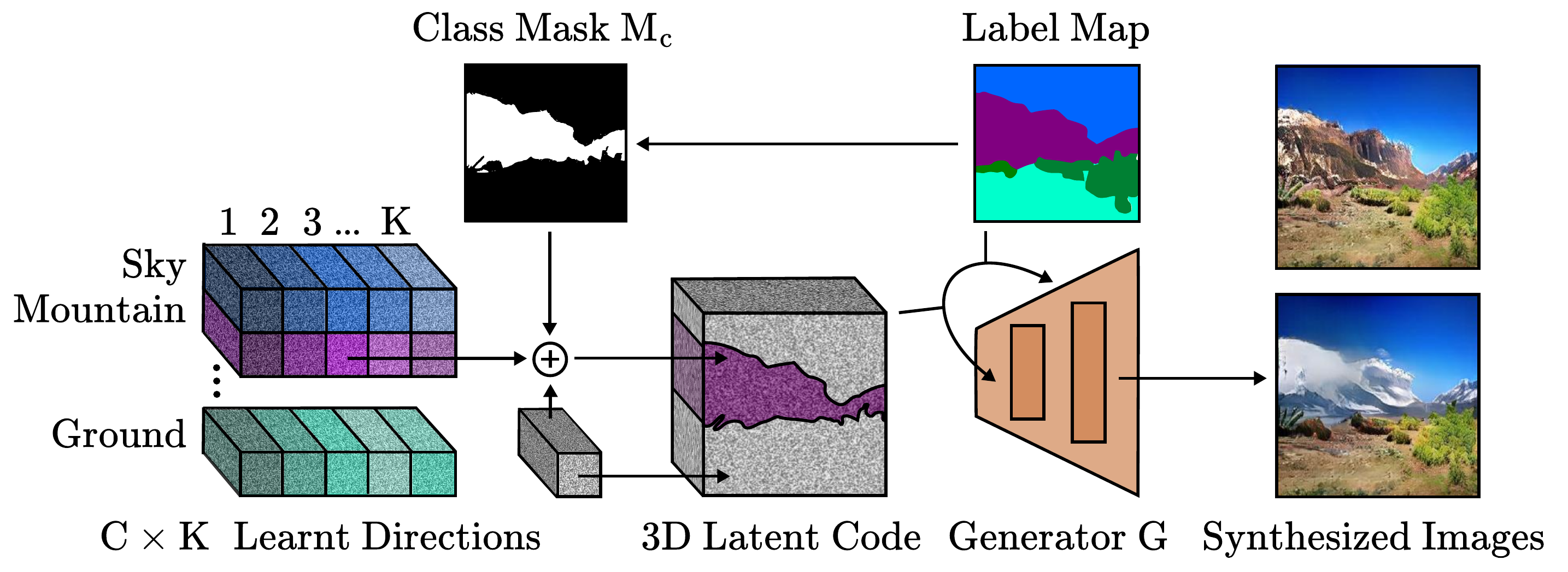} 
\par\end{centering} 
\caption{Ctrl-SIS provides a set of $K$ class-specific latent directions which control the appearance of $C$ semantic classes (shown left). To alter the appearance of class $c$, a class-specific latent direction is added to the input 3D latent code $z$ of the pretrained generator $G$ in the label map area corresponding to class $c$ ($M_c$).}
\label{fig:method} 
\vspace{-0.5em}
\end{figure*}

%% file: tex/method.tex
\section{Ctrl-SIS method} \label{sec:method}

\input{fig/qual_results.tex}

The goal of this work is to discover steerable latent directions for GAN-based SIS models. Enabled by the given semantic label maps, we aim to find GAN controls specific to semantic classes, e.g. a set of latent directions for controlling the appearance of the street and another set of directions for the appearance of house facades, see~Fig.\ref{fig:teaser}. However, this task presents two major challenges.

The first challenge is that SIS GANs commonly do not provide the same image diversity as unconditional models~\cite{brock2018large,Karras2019stylegan2}, nor have region-specific latent codes.
We alleviate both problems by applying a 3D latent code injection~\cite{schonfeld2020you}, which we describe in Sec.~\ref{subsec:gan_controls}.
The second challenge is that prior GAN control discovery methods are not designed to consider label maps, 
nor to find \textit{class-specific} directions, as they are devised for unconditional GANs.
We address both aspects in Sec.~\ref{subsec:gan_control_discovery} with a simple, efficient optimisation method which we call Ctrl-SIS. 

\subsection{GAN controls for SIS models}
\label{subsec:gan_controls}

Current SIS models employ different ways to inject latent code into the generator, which affects its ability to perform class-specific edits. 
The default approach is to feed a one-dimensional latent vector as input to the generator~\cite{park2019semantic,wang2021image,liu2019learning}, resulting in no direct opportunity to perform local region-based edits of the image. Thus, in order to enable local editing in SIS, we employ the 3D latent code injection scheme from~\cite{schonfeld2020you}, adopting it to all SIS models considered in this work.
The 3D latent codes $z \in \mathbb{R}^{H \times W \times D} $ are created by replicating the original noise vector along the height $H$ and width $W$ of the label map.  
The 3D latent code allows to apply different latent vectors to different image regions (see Fig.~\ref{fig:method}).
In practice, altering the 3D latent code only for a specific image region can still affect other image areas, due to spatial correlations learnt by the generator during training.
Nevertheless, the 3D latent space provides better spatial disentanglement and, thus, improves image manipulation control for local edits compared to 1D latent codes. 
In the remainder of this paper, we assume a 3D latent space for the discovery of SIS GAN controls.

Let $G$ be a well-trained GAN generator of a SIS model.
The generator $G(z,y)$ synthesises an image given a 3D latent code $z$ and label map $y$, 
i.e. $x = G(z,y) = F(h(z,y))$, where $h = \{G_l(z,y)\}_{l \in L}$ is a chosen subset of features from intermediate layers $l \in L$ in the network $G$, and $C$ is the total number of semantic classes. 
The latent code $z$ controls the appearance of the synthetic image, while the label map $y$ specifies the scene layout.
Then an image $x$ can be globally edited by moving $z$ along a specific direction $v_k$:
\begin{equation}
x(v_k)= F(h(z,v_k,y))=G(z+\alpha v_k, y),
\label{Eq:global_edit}
\end{equation} 
where $\alpha$ controls the intensity of the change, and the latent direction $v_k$ determines the semantics of the image transformation.
Local editing of class $c$ in $x$ is done by moving $z$ along a class-specific direction $v^c_k$ only in the area of class $c$ in the label map $y$:
\begin{equation}
x(v^c_k)= F(h(z,v^c_k,y))=G(z+\alpha M_c \odot v^c_k, y),
\label{Eq:class_edit}
\end{equation} 
where $M_c=\mathbbm{1}_{[y=c]}$ is a binary mask indicating pixels in the image belonging to $c$ (see Fig. \ref{fig:method}).
We next define the task of class-specific GAN control discovery and introduce an optimization objective to find $v^c_k$ directions for any pre-trained SIS model with a spatially-aware generation process induced by 3D latent codes.

\subsection{Discovery of class-specific GAN controls}
\label{subsec:gan_control_discovery}
For the class of interest $c \in C$ we aim to find a diverse set of class-specific directions $V^c = \{v^c_0, v^c_1, ..., v^c_K\}$, $K>1$, that can meaningfully edit the appearance of class $c$ in the synthetic image $x$, such that image $x(v^c_k)$ has a visually distinct appearance of class $c$ compared to $x$, but all other classes have the same appearance as in $x$.
Based on this logic, we form an optimization objective, which consists of diversity, disentanglement and consistency loss terms:
\begin{equation}\label{eq:objective}
\min_{V^c}\ \mathcal{L}_{div} + \mathcal{L}_{dis} + \mathcal{L}_{const}\ .
\end{equation}
The diversity loss $\mathcal{L}_{div}$ encourages a set of class-specific GAN controls $V^c$ to be diverse and introduce different semantic changes to class $c$, the disentanglement loss $\mathcal{L}_{dis}$ prevents changes outside the class area, and the consistency loss $\mathcal{L}_{const}$ ensures that the semantics of an edit are consistent between different initial latent codes $z$. We next provide the mathematical formulation of these loss terms.
\input{fig/tree_removal.tex}

\myparagraph{Diversity loss.} 
Given a label map $y$ and a class of interest $c$, the diversity loss aims to ensure that the set of found latent directions $V^c$ applied to identical input latent code $z$ yields maximally different semantic visual effects, i.e. change an appearance of class $c$ in a different way. It is formulated as %
\begin{equation}
\thickmuskip=0mu
\small
\mathcal{L}_{div}\! =\! -\mathbb{E}_{(z,y)} \Big[ \sum^K_{\substack{k_{1,2}=1 \\ k_1\neq k_2}} \!M_c \cdot ||h(z,v^c_{k_1},y)-h(z,v^c_{k_2},y)||_2 \Big], 
\end{equation}
where $\Vert \cdot\Vert$ is the $L_2$ norm, 
and for the class-specific area $M_c$ the distance between the two resulting images $x(v^c_{k_1})$ and $x(v^c_{k_2})$ is maximised in the generator feature space $h$,  
ensuring semantically different directions for class $c$. Depending on the selected feature space in $G$, i.e. the subset of intermediate layers $L$ in $h = \{G_l(z,y)\}_{l \in L}$, we can find various GAN control directions which correspond to different semantics encoded in the selected feature space of $G$.

\myparagraph{Disentanglement loss.} 
The discovered latent direction $v^c_k$ for class $c$ should only affect the image area belonging to $c$ in the label map $y$ and leave the rest of the image unaffected. 
Thus, we also minimize the change for images $x(v^c_{k_1})$ and $x(v^c_{k_2})$ in the feature space $h$ in the area outside of $M_c$:
\begin{equation}
\thickmuskip=0mu
\small
\mathcal{L}_{dis} = \mathbb{E}_{(z,y)} \Big[\! \sum^K_{\substack{k_{1,2}=1 \\ k_1\neq k_2}} \! (1\!-\!M_c) \cdot ||h(z,v^c_{k_1},y)\!-\!h(z,v^c_{k_2},y)||_2 \Big]. 
\end{equation}

\myparagraph{Consistency loss.} 
Identical GAN control directions should cause consistent semantic edits of class $c$ for different input latent codes and the same label map $y$ given to the generator.
Therefore, for every found direction $v^c_k$ we minimize the feature space distance between two images generated with $z_1$ and $z_2$ in the class-specific area $M_c$ : 
\begin{equation}
\thickmuskip=0mu
\small
\mathcal{L}_{const} = \mathbb{E}_{(z,y)} \Big[ \sum_{k=1}^{K} M_c \cdot ||h(z_1,v^c_k,y)-h(z_2,v^c_k,y)||_2 \Big]. 
\end{equation}

Note that the directions in $V^c$ are the only parameters to be optimized; the weights of the pre-trained image generator $G(z,y)$ are kept frozen. The parameters are optimized by iterating over batches of label maps in the training set and minimizing the objective for selected classes at every step. 
During optimization, the directions $v^c_k$ are normalized along the channel dimension to unit length 1 and subsequently scaled by $\alpha$, sampled from the interval $[-n;n]$, where $ n= \mathop{\mathbb{E}} [||z||_2] $ is the average norm of the latent code along the channel dimension. This ensures the latent edits are neither too small nor too extreme.

%% file: fig/qual_results.tex
\begin{figure*}[t]
\begin{centering}
\setlength{\tabcolsep}{0.1em}
\renewcommand{\arraystretch}{0}
\par\end{centering}
\begin{centering}
\vspace{-1em}
\hfill{}
\begin{tabular}{c@{\hskip 0.05in}c@{\hskip 0.05in}c@{\hskip 0.05in}c@{\hskip 0.15in}c@{\hskip 0.05in}c@{\hskip 0.05in}c@{\hskip 0.05in}c}

    \multirow{1}{*}{ \rotatebox{90}{  \blue{Tree} \hspace{4.0em}  \blue{Face} \hspace{3.7em} \blue{Window} \hspace{-4.5em} }}  
    
    & 
    \includegraphics[width=0.14\textwidth, height=0.09\textheight]{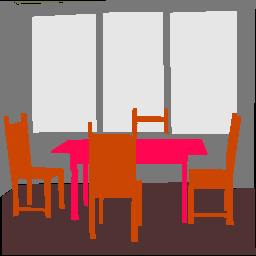} & 
    \includegraphics[width=0.14\textwidth, height=0.09\textheight]{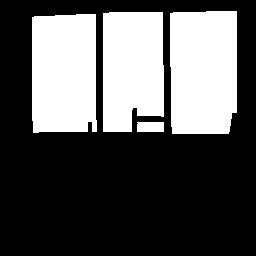} & 
    \includegraphics[width=0.14\textwidth, height=0.09\textheight]{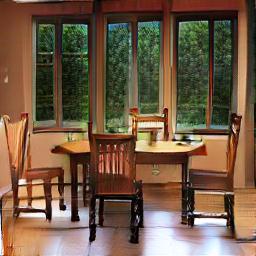} & 
    \includegraphics[width=0.14\textwidth, height=0.09\textheight]{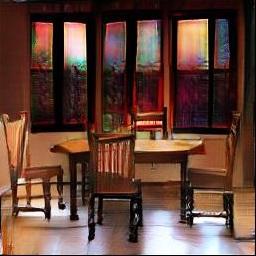} & 
    \includegraphics[width=0.14\textwidth, height=0.09\textheight]{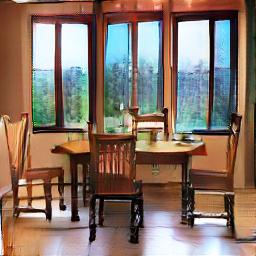} & 
    \includegraphics[width=0.14\textwidth, height=0.09\textheight]{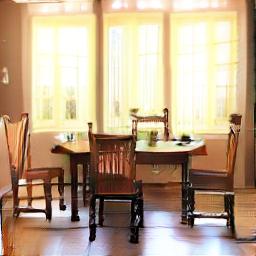}  \\

   &
    \includegraphics[width=0.14\textwidth, height=0.10\textheight]{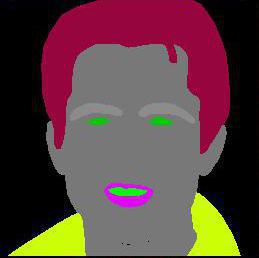} & 
    \includegraphics[width=0.14\textwidth, height=0.10\textheight]{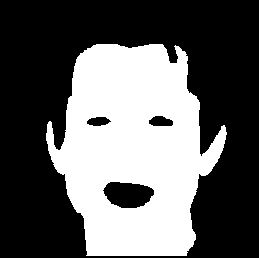} & 
    \includegraphics[width=0.14\textwidth, height=0.10\textheight]{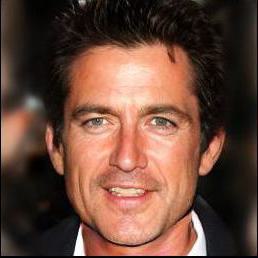} & 
    \includegraphics[width=0.14\textwidth, height=0.10\textheight]{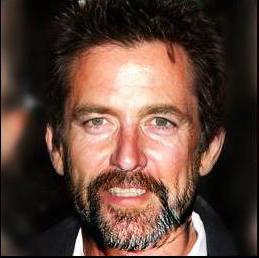} &  
    \includegraphics[width=0.14\textwidth, height=0.10\textheight]{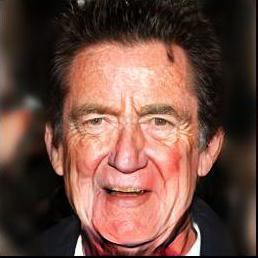} &  
    \includegraphics[width=0.14\textwidth, height=0.10\textheight]{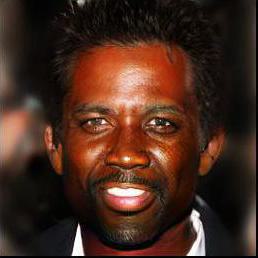}  \\
    
  & 
    \includegraphics[width=0.14\textwidth, height=0.09\textheight]{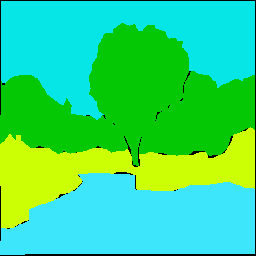} & 
    \includegraphics[width=0.14\textwidth, height=0.09\textheight]{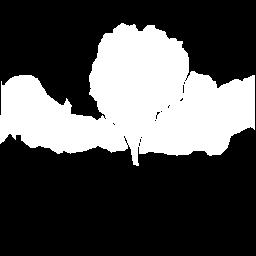} & 
    \includegraphics[width=0.14\textwidth, height=0.09\textheight]{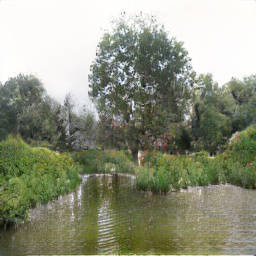} & 
    \includegraphics[width=0.14\textwidth, height=0.09\textheight]{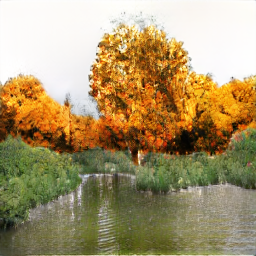} &  
    \includegraphics[width=0.14\textwidth, height=0.09\textheight]{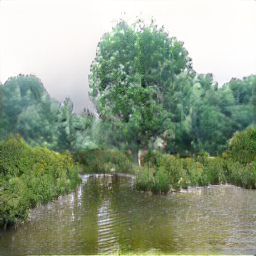} & 
    \includegraphics[width=0.14\textwidth, height=0.09\textheight]{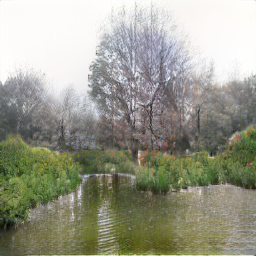}  \\

    &  \normalsize{Label map} & \normalsize{Class} & \normalsize{Original} & \normalsize{k=1}& \normalsize{k=2} & \normalsize{k=3}

\end{tabular}
\hfill{}

\par\end{centering} 
\vspace{-0.5em}
\caption{Examples of directions discovered by Ctrl-SIS for various classes, such as different views of a window, face appearances or tree leafage for different seasons of the year. The directions give insight into the concepts that the pretrained SIS model is able to represent.} %
\label{fig:qual_results} 
\vspace{-0.5em}
\end{figure*}

%% file: fig/tree_removal.tex
\begin{figure*}[t]
	\begin{centering}
		\setlength{\tabcolsep}{0.1em}
		\renewcommand{\arraystretch}{0}
		\par\end{centering}
	\begin{centering}
		\vspace{-1em}
		\hfill{}
		
		\begin{tabular}{c@{\hskip 0.042in} c@{\hskip 0.042in} c@{\hskip 0.042in} c@{\hskip 0.092in} c@{\hskip 0.042in} c@{\hskip 0.042in} c@{\hskip 0.042in} c}

			\multirow{1}{*}{ \rotatebox{90}{ \small  Edited \emph{lamp}  \hspace{1.6em} Original \hspace{-4.0em} }}  & 
			\includegraphics[width=0.14\textwidth, height=0.09\textheight]{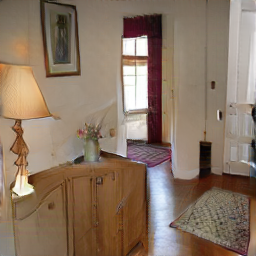} & %
			\includegraphics[width=0.14\textwidth, height=0.09\textheight]{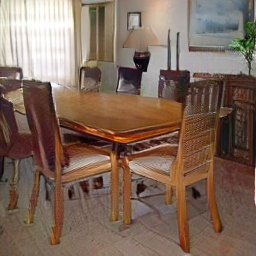} & %
			\includegraphics[width=0.14\textwidth, height=0.09\textheight]{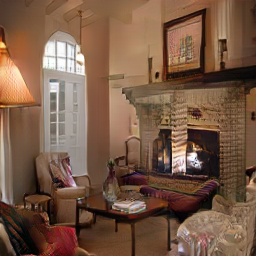} & 

			\multirow{1}{*}{ \rotatebox{90}{ \small  Edited \emph{tree}  \hspace{1.8em} Original \hspace{-4.0em} }}  & 
			\includegraphics[width=0.14\textwidth, height=0.09\textheight]{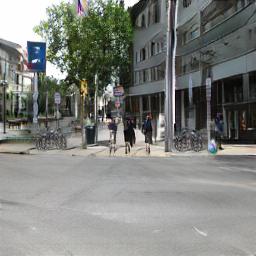} & 
			\includegraphics[width=0.14\textwidth, height=0.09\textheight]{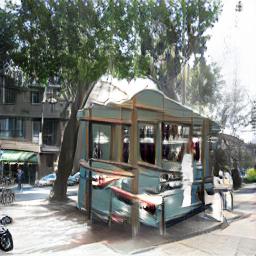} &  
			\includegraphics[width=0.14\textwidth, height=0.09\textheight]{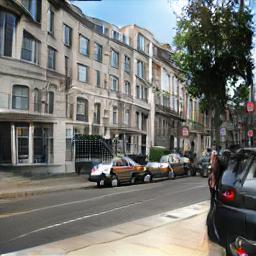} 

			\\ 

			& 
			\includegraphics[width=0.14\textwidth, height=0.09\textheight]{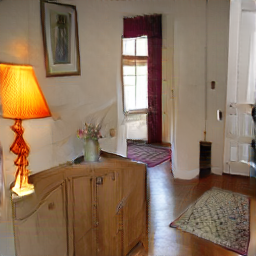} & %
			\includegraphics[width=0.14\textwidth, height=0.09\textheight]{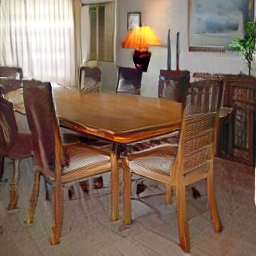} & %
			\includegraphics[width=0.14\textwidth, height=0.09\textheight]{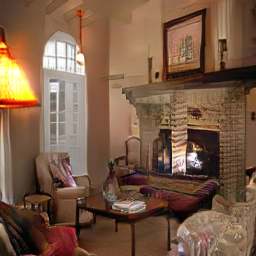} & 		
			 & 
			\includegraphics[width=0.14\textwidth, height=0.09\textheight]{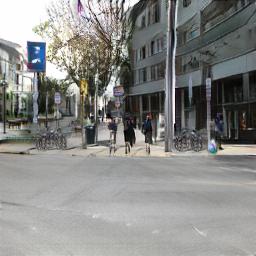} & 
			\includegraphics[width=0.14\textwidth, height=0.09\textheight]{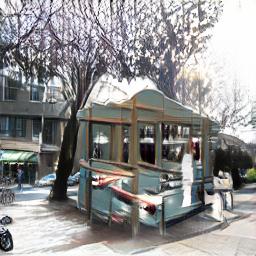} &  
			\includegraphics[width=0.14\textwidth, height=0.09\textheight]{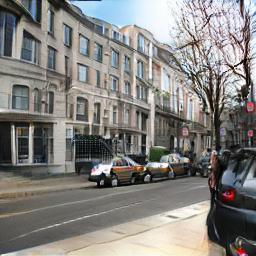}

			\\ \vspace{-0.5em}
			\\
			
			\multirow{1}{*}{ \rotatebox{90}{\small Edited \emph{road}  \hspace{1.5em}  Original \hspace{-4.0em} }}   &
			\includegraphics[width=0.14\textwidth, height=0.09\textheight]{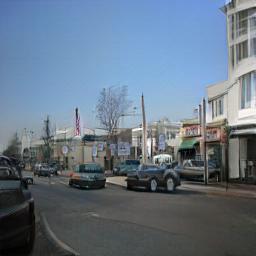} & 
			\includegraphics[width=0.14\textwidth, height=0.09\textheight]{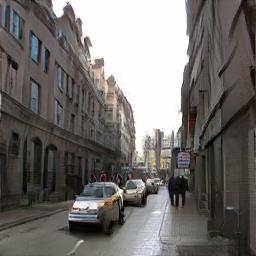} & 
			\includegraphics[width=0.14\textwidth, height=0.09\textheight]{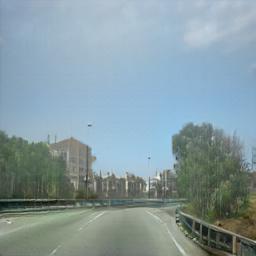} & 
			
			\multirow{1}{*}{ \rotatebox{90}{\small Edited \emph{mount.}  \hspace{1.5em}  Original \hspace{-4.0em} }}   &
			\includegraphics[width=0.14\textwidth, height=0.09\textheight]{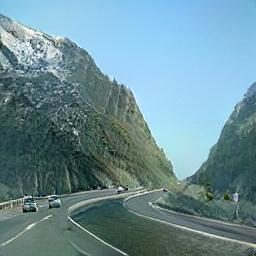} & 
			\includegraphics[width=0.14\textwidth, height=0.09\textheight]{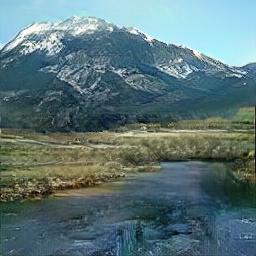} & 
			\includegraphics[width=0.14\textwidth, height=0.09\textheight]{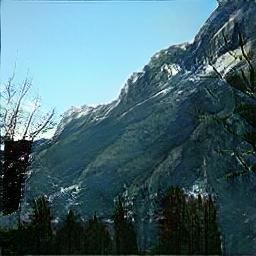} 
			\\ 
			
			& 
			\includegraphics[width=0.14\textwidth, height=0.09\textheight]{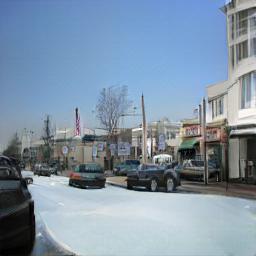} & 
			\includegraphics[width=0.14\textwidth, height=0.09\textheight]{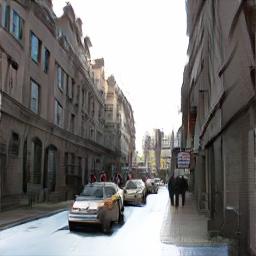} & 
			\includegraphics[width=0.14\textwidth, height=0.09\textheight]{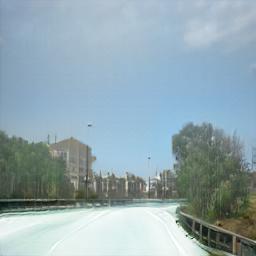} &
			
			 &
			\includegraphics[width=0.14\textwidth, height=0.09\textheight]{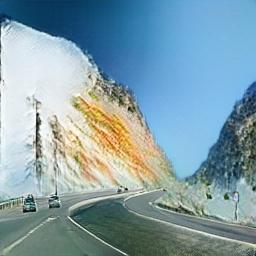} & 
			\includegraphics[width=0.14\textwidth, height=0.09\textheight]{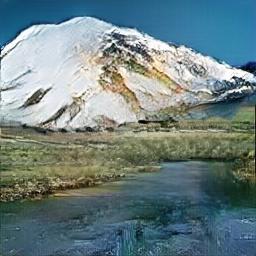} & 
			\includegraphics[width=0.14\textwidth, height=0.09\textheight]{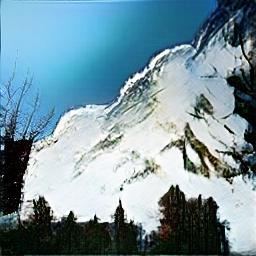} 
			
			\\ \vspace{-0.5em}
		\\ 	

			\multirow{1}{*}{ \rotatebox{90}{ \small  Edited \emph{face}  \hspace{1.8em} Original \hspace{-4.0em} }}  & 
			\includegraphics[width=0.14\textwidth, height=0.10\textheight]{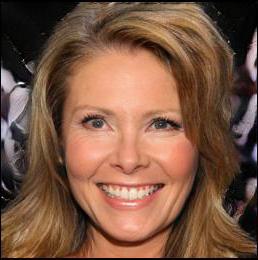} & 
			\includegraphics[width=0.14\textwidth, height=0.10\textheight]{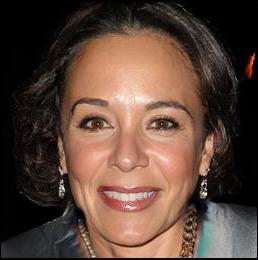} &  
			\includegraphics[width=0.14\textwidth, height=0.10\textheight]{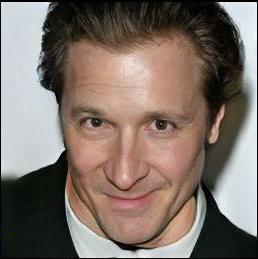} &  

			\multirow{1}{*}{ \rotatebox{90}{ \small  Edited \emph{hair}  \hspace{1.8em} Original \hspace{-4.0em} }}  & 
			\includegraphics[width=0.14\textwidth, height=0.10\textheight]{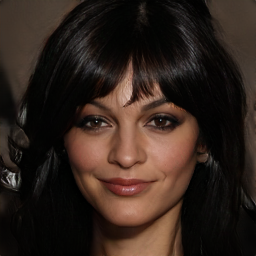} & 
			\includegraphics[width=0.14\textwidth, height=0.10\textheight]{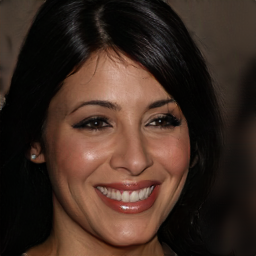} &  
			\includegraphics[width=0.14\textwidth, height=0.10\textheight]{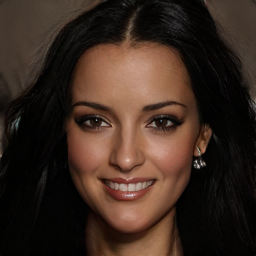} 
			
			\\

			 & 
			\includegraphics[width=0.14\textwidth, height=0.10\textheight]{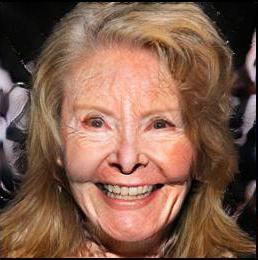} & 
			\includegraphics[width=0.14\textwidth, height=0.10\textheight]{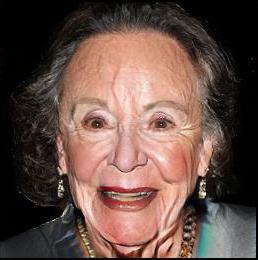} &  
			\includegraphics[width=0.14\textwidth, height=0.10\textheight]{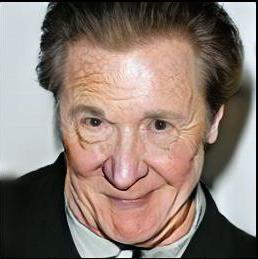} &  

			 & 
			\includegraphics[width=0.14\textwidth, height=0.10\textheight]{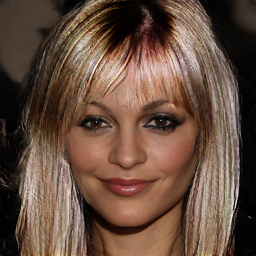} & 
			\includegraphics[width=0.14\textwidth, height=0.10\textheight]{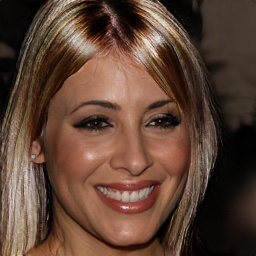} &  
			\includegraphics[width=0.14\textwidth, height=0.10\textheight]{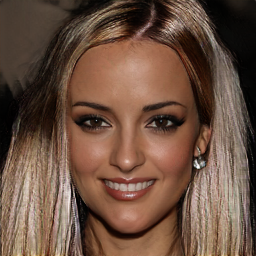} 
			
		\end{tabular}
		
		\par\end{centering}
	\vspace{-0.5em}
	\caption{Interpretable latent directions learnt by Ctrl-SIS for various classes. Each triplet is edited with an identical direction. Class-specific edits, such as aging, snowy streets or bald trees, are highly consistent across different label maps and initial latent codes. 
	}
 \vspace{-0.5em}
	\label{fig:tree_removal}
\end{figure*}

%% file: tex/experiments.tex
\section{Experiments}\label{sec:experiments}

\input{tables/Table1_coco.tex}

\subsection{Experimental setup}\label{subsec:exp_setup}
\myparagraph{Datasets.} 
We use three challenging datasets: CelebAMask-HQ~\cite{lee2020maskgan}, ADE20K~\cite{zhou2017scene}, and COCO-Stuff~\cite{caesar2018coco}.
CelebAMask-HQ consists of 30k face images. ADE20K and COCO-Stuff contain 20k and 164k images of indoor and outdoor scenes, and are used for the main experiments. 
 
\myparagraph{SIS models.} 
We consider three pre-trained GANs for SIS: SC-GAN~\cite{wang2021image}, SPADE~\cite{park2019semantic} and OASIS~\cite{schonfeld2020you}, using the code provided by the authors 
\footnote{Code for SIS models: \href{https://github.com/NVlabs/SPADE}{SPADE}, \href{https://github.com/dvlab-research/SCGAN}{SC-GAN}, \href{https://github.com/boschresearch/OASIS}{OASIS}}.
We additionally implement 3D latent codes for SPADE and SC-GAN, which do not originally support it, enabling local image editing for them. 

\myparagraph{GAN control methods.} 
Ctrl-SIS is compared against the two related latent discovery methods GANSpace~\cite{harkonen2020ganspace} and SeFa~\cite{shen2021closed}, using the authors' code
\footnote{Code for GAN control methods: \href{https://github.com/harskish/ganspace}{GANSpace}, \href{https://github.com/genforce/sefa}{SeFa}}.
Following GANSpace-StyleGAN2~\cite{harkonen2020ganspace} and SeFA-StyleGAN2~\cite{shen2021closed}, we train all latent direction methods on features extracted from the normalization layers of each ResNet block in the generator. 

\myparagraph{Training details.}
Ctrl-SIS is trained with a batch size of 16 on a NVIDIAv100 GPU, using the AdamW optimizer~\cite{loshchilov2017decoupled} and a learning rate of 1e-3. We train for 20 epochs on ADE20K, and 5 epochs on COCO-Stuff and CelebAMask-HQ, using $K=5$. Finding class-specific directions with Ctrl-SIS takes $\sim$1h. 
For evaluation we scale the directions with $\alpha$ sampled in $[-n;n]$ (see Sec.~\ref{subsec:gan_control_discovery}), to ensure that the direction magnitude is in the same range as the average latent code. By scaling the magnitudes of latent directions from all methods in the same way, we ensure that the effect of the edit only depends on the learnt direction. For GANSpace and SeFa, we pick the directions belonging to the first $K$ components, as they cause the largest variations.

\subsection{Evaluation}\label{subsec:exp_eval}

\myparagraph{Image quality evaluation.}
Following \cite{isola2017image,park2019semantic,schonfeld2020you}, we monitor the visual quality of images generated with class-specific edits using FID~\cite{heusel2017gans} and mIoU metrics. 
FID is known to be well aligned with human judgement of image quality. 
mIoU assesses the alignment of images with ground truth label maps, calculated via a pre-trained semantic segmentation network. 
We use UperNet101~\cite{xiao2018unified} for ADE20K and DeepLabV2 \cite{chen2014semantic} for COCO-Stuff. 
In addition, we employ the precision and recall metrics of~\cite{kynkaanniemi2019improved}, which
correlate with image quality and diversity, respectively.

\myparagraph{Evaluation of class-specific GAN controls.} %
Prior GAN control methods were mostly evaluated by subjective visual inspection~\cite{harkonen2020ganspace,shen2021closed,yuksel2021latentclr}. 
Consequently, it was challenging to assess the important properties of GAN control methods. First, how many unique and visually distinct latent directions are found. 
Second, how general are the directions, i.e. if they consistently invoke the same semantic change in all images. Third, in the case of SIS, local class-specific edits must not affect the rest of the image. To quantitatively assess these properties, we introduce the following metrics. 

To measure how unique and distinct the found directions are, we introduce the \textit{mean control diversity} (mCD). mCD measures the mean pairwise LPIPS~\cite{zhang2018perceptual} distance between edited images generated with the same label map. We generate 10 global edits of each synthetic test set image by applying a randomly selected class-specific direction to each class area in the label map. The pairwise LPIPS distance is then averaged over all images and 5 initial input latent codes. 
We also measure a local version mCD$_l$, where only a single class is locally edited at a time. In this case, we compute the average masked LPIPS distance inside the class-specific area of a single class for all pairs of the $K$ class-specific directions. The final mCD$_l$ score is the average of the mean per-class scores. A high mCD$_l$ score implies that each discovered latent direction changes the class appearance in a uniquely different way.

To evaluate how much class-specific edits affect the areas outside of the target class area, we introduce the \textit{mean outside class diversity} (mOD). mOD is computed similarly to mCD$_l$, except that we use the masked LPIPS distance in the image area outside of the target class area. Ideally, mOD is very low, as we want the latent direction to alter only the class-specific region. Both mCD and mOD scores are inspired by the metrics proposed by~\cite{zhu2020semantically}, see Sec. B in the supplementary for more details.
 
We introduce the \textit{mean control consistency} (mCC) score to measure to which extent latent directions invoke the same interpretable edit in all images. The mCC is computed by editing each class in an image with a randomly picked class-specific direction. In this case, the score of an image is the mean pairwise LPIPS distance when different initial latent codes are applied while the class-specific latent direction is kept the same. The final mCC score is averaged over all label maps in the test set. Same as for mCD$_l$, for the local mCC$_l$ score, the edits are only applied to one class and the masked LPIPS distance is used. 
The lower mCC$_l$ the more consistent are the class-specific edits for different initial latent codes.
A detailed further description of the metrics is given in Sec. B in the supplementary. With the introduced metrics, a more unbiased and systematic comparison of GAN control discovery methods for SIS is possible.

\myparagraph{Human evaluation.} %
We also conduct a human evaluation of the learnt latent directions.
To this end, we employ the SHE score from \cite{zhu2020semantically} and introduce a Human Diversity Rank (HDR) metric. %
For SHE, participants are shown two images edited only in the corresponding class area by applying the learnt class-specific latent direction. The final SHE score is the percentage of image pairs that the participants judge to be semantically different in the area of only one class.
For HDR, participants are shown rows of locally edited images from four different methods (Random, SeFa, GANSpace, Ctrl-SIS) 
, as in Fig. C in the supplementary material but in a randomized order.
The task is to rank the methods by their diversity. The final HDR score is an average rank (range 1 to 4) assigned to a GAN control discovery method. Each participant is provided with 50 questions and unlimited answering time for both scores.

\input{fig/celeba_combi.tex}
\input{fig/ade_combi}

\subsection{Main results} \label{subsec:exp_res}

We compare Ctrl-SIS, GANSpace and SeFa on global and local image editing. While local edits target a single class per image, global edits combine all class-specific edits within an image. 
The local edits show that the found directions encode semantic meaning, such aging faces, covering mountains in snow or turning on lamps (see Fig. \ref{fig:qual_results} and \ref{fig:tree_removal}).
Global edits, 
change the whole image globally and are the result of combining all class edits in one image.
In addition, we compare all methods to the performance of randomly sampled directions ("Random") as well as to the performance on unedited images ("Baseline").

\input{tables/alternative_metrics.tex} %

\input{tables/table_3_small_reshaped}

As seen from Table \ref{table:1_coco}, Ctrl-SIS achieves improved diversity by at least a factor of two, e.g.,  mCD of 0.26 vs. 0.12 of SeFa on ADE20K. 
The diversity of GANSpace and SeFa is lower (see numbers in red in Table \ref{table:1_coco}) or close to random directions.
Neither of these methods are designed to find class-specific directions. Yet, they still capture class-agnostic variations in the data, leading to directions that are closer to the mean of the image distribution and thus slightly better FID and mIoU.
All methods exhibit similar consistency (mCC), with Ctrl-SIS and SeFa performing best on ADE20K, and SeFa and GANSpace on COCO-Stuff. 
The consistency of Ctrl-SIS is demonstrated in Fig. \ref{fig:tree_removal}, e.g, where the learnt latent direction consistently defoliates trees or covers streets in snow.
Similar to the consistency, the disentanglement (mOD) is strong for all methods, due to the spatially disentangled 3D latent space of the OASIS model. 

Due to the higher diversity of edited images, FID increases slightly for Ctrl-SIS compared to the baseline of unedited images (see Table \ref{table:1_coco}). 
Since FID measures the overlap between the real and synthetic image distributions, images with weaker edits are closer to the original data. 
This can be visually confirmed in Fig. C in the supplementary material, 
where edits are shown side-by-side for all methods. Since SeFa and GANSpace only minimally change the class, their FID is close to the FID of unedited images (see Baseline in Table \ref{table:1_coco}). Likewise, mIoU of images edited with Ctrl-SIS decreases, as the edited images move away from the mean mode of the synthetic image distribution. In contrast, for SeFa and GANSpace FID and mIoU are slightly better with respect to the baseline, while their diversity (mCC) is comparable to random directions. This observation suggests that SeFa and GANSpace images are closer to typical samples of the test set, while Ctrl-SIS learns more distinct directions.

We perform alternative evaluations of local class-specific edits in Table~\ref{table:ctrlsis_alternative_eval}.
Ctrl-SIS shows the highest recall and diversity (${\mathrm{mCD}_l}$), and is the only method to improve both precision and recall over the OASIS baseline.  
Due to the precision-recall trade-off, SeFa and GANSpace have higher precision at the loss of recall, which is also reflected in their low ${\mathrm{mCD}_l}$ score and better FID over Baseline in Table~\ref{table:1_coco}.
Moreover, both human diversity evaluation scores (SHE and HDR) are well aligned with the diversity metric ${\mathrm{mCD}_l}$ and recall, confirming the highest diversity of Ctrl-SIS.

Next, we compare Ctrl-SIS on different SIS models. 
Table \ref{table:sis_models} shows that Ctrl-SIS improves diversity strongly for local edits across all tested SIS models. 
SPADE naturally suffers from lower sensitivity to input latent code due to the strong regularization effect of its perceptual loss, as shown in~\cite{schonfeld2020you}. While OASIS is trained without a perceptual loss, and SC-GAN uses a more powerful layer-wise conditioning strategy, leading to more diversity.
Similar to Table \ref{table:1_coco}, the diversity of GANSpace and SeFa is comparable to random directions. 
In other words, the directions that SeFa and GANSpace find differ just as much from each other as a set of randomly chosen directions.
In contrast, the directions of Ctrl-SIS embody distinct appearances that are unlikely to appear in a set of random directions. An extended version of Table \ref{table:sis_models} with global metrics, FID and mIoU can be found in Table C in the supplementary material.

\myparagraph{Compositionality.}  
Individual class-specific latent directions can be combined. For example, Fig. \ref{fig:celeba_directions} shows that directions corresponding to "age" and "beard" can be combined into "old and bearded".
Further, the latent directions found by Ctrl-SIS depend on the subset of feature layers $G_l(z,y)$ of the SIS generator $G$ chosen for optimization (see Sec.~\ref{sec:method}).  
Fig. \ref{fig:qual_compose} highlights latent directions that were discovered by optimizing over layers from different blocks of the generator. 
For example, the set Norm 4 in Fig. \ref{fig:qual_compose} minimizes the loss over the first convolution in all conditional normalization layers~\cite{park2019semantic} within the fourth ResNet block. While for the set Res 1 we minimized the loss for the final output features of the first ResNet block.
We observe that the directions for Res 1 and 2 differ strongly in the internal structure of semantic classes, while Norm 3 and 4 encode changes in color.
Interestingly, latent directions can be combined when synthesising images, by injecting different directions in different layers. In the last column of Fig. \ref{fig:qual_compose}, a direction of an early ResNet block is injected into the first four blocks of the SIS model, while directions from the conditional normalization layers of a late ResNet blocks are injected from block five onwards. As the former directions encode structure, and the latter encodes colors, the resulting image combines both aspects.

\input{tables/loss_ablation.tex}

\myparagraph{Ablation.} 
Table \ref{table:loss_ablation} presents an ablation on the proposed objective, using OASIS on the ADE20K dataset. Without the diversity term in our objective, the diversity decreases. Likewise, without the consistency or disentanglement term, consistency and disentanglement numbers worsen, respectively (see numbers in red). Further, the disentanglement term helps to improve synthesis and segmentation quality (see FID and mIoU in red), by helping to restrict the area affected by the edit only to the selected class area.

%% file: tables/Table1_coco.tex
\setlength{\tabcolsep}{0.5em}
\renewcommand{\arraystretch}{0.9}

\begin{table*}[t] 
	\centering
	\vspace{-1em}	
		
		\begin{tabular}{l|ccccc|ccccc}
			\multirow{2}{*}{ Method} 
			& \multicolumn{5}{c|}{ADE20K} 
			& \multicolumn{5}{c}{COCO-Stuff} 
			\tabularnewline

			& \normalsize ${\mathrm{mCD}}\uparrow$ 
			& \normalsize ${\mathrm{mCC}}\downarrow$                 
			& \normalsize ${\mathrm{mOD}}\downarrow$ 
			& \normalsize FID$\downarrow$ 
			& \normalsize mIoU$\uparrow$ 
			& \normalsize ${\mathrm{mCD}}\uparrow$ 
			& \normalsize ${\mathrm{mCC}}\downarrow$ 
			& \normalsize ${\mathrm{mOD}}\downarrow$ 
			& \normalsize FID$\downarrow$ 
			& \normalsize mIoU$\uparrow$
			\tabularnewline 
			
			\hline \hline
			
			Baseline  
			& -
			& -
			& -  
			& 28.6      
			& 52.2 
			& -
			& -   
			& - 
			& \textbf{17.1}  
			&  42.4
			\tabularnewline

			\hline
			
			Random    
			& 0.11     
			& 0.30     
			& \textbf{0.01} 
			& 31.3      
			& 49.4     
			& 0.16     
			& 0.07 
			& \textbf{0.00}	
			& 17.6   
			& 42.3
			\tabularnewline

			GANSpace 
			& \emphtable{0.09}  %
			& 0.29     
			& \textbf{0.01} 
			& \textbf{28.1}
			& \textbf{53.3}       
			& \emphtable{0.15}  %
			& \textbf{0.06}   
			& \textbf{0.00}  
			& 17.2   
			& 42.1
			\tabularnewline

			SeFa   
			& 0.12     
			& \textbf{0.28}       
			& \textbf{0.01} 
			& \textbf{28.1}
			& 53.2     
			& \emphtable{0.15} %
			& \textbf{0.06} 
			& \textbf{0.00} 
			& \textbf{17.1}    
			& \textbf{43.8}   
			\tabularnewline

			\textbf{Ctrl-SIS} 
			& \textbf{0.26}       
			& \textbf{0.28}       
			& \textbf{0.01} 
			& 30.9      
			& 48.9     
			& \textbf{0.30}       
			& 0.07		
			& 0.01				
			& 21.1     
			& 43.6
			\tabularnewline

		\end{tabular}%
		\vspace{-0.5em}
		\caption{Evaluation of OASIS GAN controls on ADE20K and COCO-Stuff. Red numbers indicate lower-than-random performance. Ctrl-SIS discovers significantly more diverse directions (mCD) without sacrificing consistency (mCC) or spatial disentanglement (mOD).\label{table:1_coco}}
		\vspace{-0.5em}
\end{table*}

%% file: fig/celeba_combi.tex
\begin{figure*}[t]
\vspace{-1em}
    \centering
    	\setlength{\tabcolsep}{0.1em}
		\renewcommand{\arraystretch}{0.9}
\begin{tabular}{@{}c@{\hskip 0.05in}c@{\hskip 0.15in}c@{\hskip 0.05in}c@{\hskip 0.05in}c@{\hskip 0.15in}c@{\hskip 0.05in}c@{\hskip 0.05in}c@{}}
        Label map & Original & $k_1$ & $k_2$ & $k_3$ & $k_1$ + $k_2$ & $k_2$ + $k_3$ & $k_1$ + $k_3$  \\
        \includegraphics[width=0.115\textwidth]{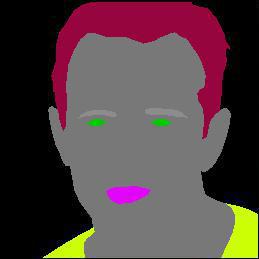} &
        \includegraphics[width=0.115\textwidth]{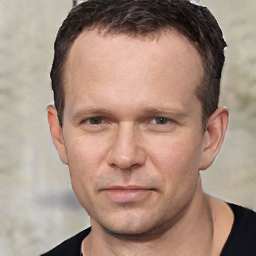} &
        \includegraphics[width=0.115\textwidth]{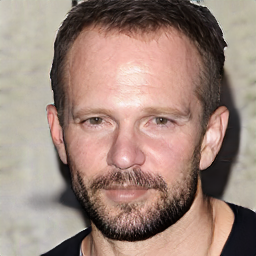} &
        \includegraphics[width=0.115\textwidth]{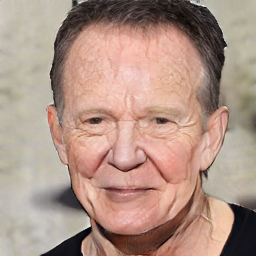} &
        \includegraphics[width=0.115\textwidth]{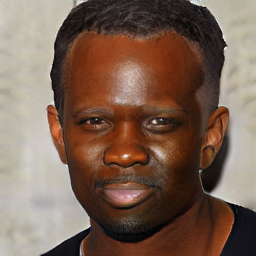} &
        \includegraphics[width=0.115\textwidth]{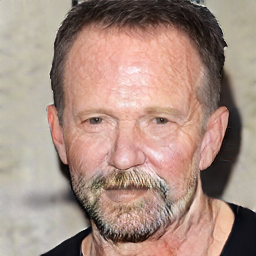} &
        \includegraphics[width=0.115\textwidth]{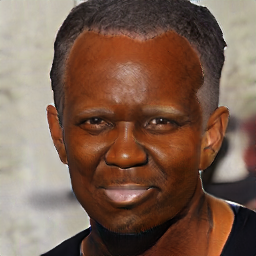} &
        \includegraphics[width=0.115\textwidth]{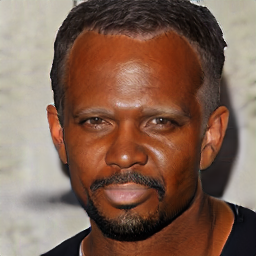} 
    \end{tabular}
\vspace{-0.5em}
    \caption{Combinations of directions $k_1$, $k_2$ and $k_3$ found for the \textit{face} class (skin, neck, nose, ears) in the CelebAMask-HQ dataset. Different directions can be added in the latent space to combine their semantics, e.g., \textit{k1 (beard)} and \textit{k2 (age)} yield an old bearded person. 
    }
	\label{fig:celeba_directions}
\end{figure*}

%% file: fig/ade_combi.tex
\begin{figure*}[t]
\begin{centering}
\setlength{\tabcolsep}{0.1em}
\vspace{-0.5em}
\begin{tabular}{@{}c@{\hskip 0.05in}c@{\hskip 0.15in}c@{\hskip 0.05in}c@{\hskip 0.15in}c@{\hskip 0.05in}c@{\hskip 0.15in}c@{}}
    \includegraphics[width=0.13\textwidth, height=0.08\textheight]{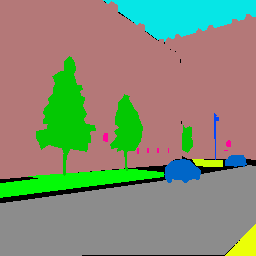} &
    \includegraphics[width=0.13\textwidth, height=0.08\textheight]{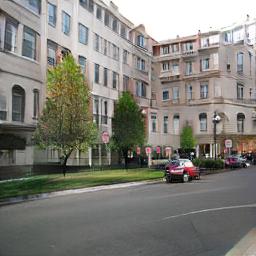}  &
     \includegraphics[width=0.13\textwidth, height=0.08\textheight]{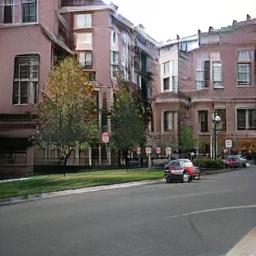} &
     \includegraphics[width=0.13\textwidth, height=0.08\textheight]{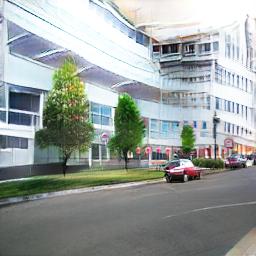} &
     \includegraphics[width=0.13\textwidth, height=0.08\textheight]{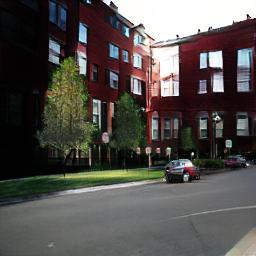} &
     \includegraphics[width=0.13\textwidth, height=0.08\textheight]{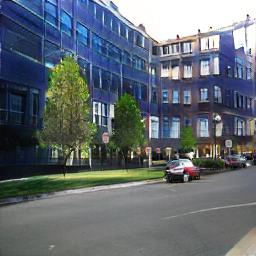}&
     \includegraphics[width=0.13\textwidth, height=0.08\textheight]{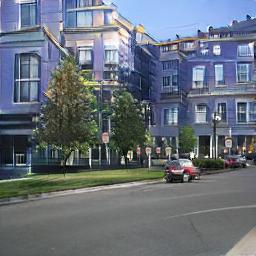} 
        \\
     \normalsize{Label map} & \normalsize{Original} & \normalsize{Res 1} & \normalsize{Res 2} & \normalsize{Norm 3} & \normalsize{Norm 4} & \normalsize{Res 1 + Norm 4} \\
    \end{tabular} 
\vspace{-0.5em}
\par\end{centering} 
\caption{Optimizing Ctrl-SIS on the output features of ResNet blocks leads to an emphasis on structure (Res 1 and 2), while late normalization layers focus more on color (Norm 3 and 4). Directions from different layers can be combined (see last column).}
\label{fig:qual_compose} 
\end{figure*}

%% file: tables/alternative_metrics.tex
\setlength{\tabcolsep}{0.2em}
\renewcommand{\arraystretch}{0.9}

	\begin{table}[t] 
		
	\vspace{-0.5em}
		
		\centering
		\begin{tabular}{l|ccc|cc} 
			
			& 	
			
			\multirow{2}{*}{\normalsize  ${\mathrm{mCD}_l}\uparrow$ 		 }   	& %

			\multirow{2}{*}{\normalsize Precision$\uparrow$} 		&
			\multirow{2}{*}{\normalsize Recall$\uparrow$}   	&
			\multicolumn{2}{c}{\normalsize Human eval.} 	\\

			& 
			&  
			&
			& 
			\normalsize SHE$\uparrow$ & \normalsize HDR$\downarrow$ \\
			
			\hline \hline

			Baseline  & - & 0.84 & 0.63 & - & -
			\tabularnewline
			\hline
			
			Random              & 0.04 & 0.82 & 0.62  & 32.9   & 2.62 
			\tabularnewline
			
			GANSpace             & 0.03 & \textbf{0.87} & 0.61  & 29.1   & 3.43   
			
			\tabularnewline
			
			SeFa                & 0.05 & \textbf{0.87} & 0.62  & 30.3   & 2.88  
			\tabularnewline
			
			\textbf{Ctrl-SIS}   & \textbf{0.12} &  0.85 & \textbf{0.64}  & \textbf{60.7}   & \textbf{1.07}   
			\tabularnewline
		\end{tabular}
		\vspace{-0.5em}
		\caption{Evaluation of local class-specific image edits on ADE20K with OASIS. Ctrl-SIS shows higher diversity of latent directions ($\mathrm{mCD}_l$, SHE, HDR, Recall) while retaining good image quality (Precision).}\label{table:ctrlsis_alternative_eval} 
		 \vspace{-0.5em}
	\end{table}

%% file: tables/table_3_small_reshaped.tex
\setlength{\tabcolsep}{0.5em}
\renewcommand{\arraystretch}{0.9}

\begin{table}[t]
	\vspace{-0.5em}
	
	\centering
	\begin{tabular}{l|c|c|c|c}
	Model & Random & GANSpace & SeFa & Ctrl-SIS \\
	\hline \hline
	OASIS & 0.04 & 0.03 & 0.05 & \textbf{0.12} \\
	SC-GAN & 0.05 & 0.06 & 0.06 & \textbf{0.18} \\
	SPADE & 0.05 & 0.08 & 0.06 & \textbf{0.09}

	\end{tabular}
	\vspace{-0.5em}
	\caption{Comparison of $\mathrm{mCD}_l$ for GAN control methods across SIS models on ADE20K. Ctrl-SIS yields more diverse latent directions independent of the pretrained SIS model, while other methods find directions comparable to random sampling. 
	\label{table:sis_models}} 
	\vspace{-0.5em}
\end{table}

%% file: tables/loss_ablation.tex
\setlength{\tabcolsep}{0.3em}
\renewcommand{\arraystretch}{0.9}
	\begin{table}[t]
			\vspace{-0.5em}
		\centering
		\begin{tabular}{l|ccccc}
			Method
		& \normalsize ${\mathrm{mCD}}$ $\uparrow$ 
		& \normalsize ${\mathrm{mCC}}$ $\downarrow$                 
		& \normalsize ${\mathrm{mOD}}$ $\downarrow$    
		& \normalsize FID $\downarrow$                     
		& \normalsize mIoU $\uparrow$ 
		\tabularnewline \hline \hline
			\textbf{Ctrl-SIS}                    & 0.26           & \textbf{0.28}      & \textbf{0.01}               & 30.9                                                                 & 48.9                                            \tabularnewline \hline
			No $\mathcal{L}_{div}$              & \emphtable{0.24}      & \textbf{0.28}     & \textbf{0.01}                     & \textbf{30.5}                                                                  & \textbf{49.4}                                              \tabularnewline 
			
			No $\mathcal{L}_{const}$             & 0.26      & \emphtable{0.29}   & \textbf{0.01}                          & 30.9                                                                   & 48.7                                                              \tabularnewline

			No $\mathcal{L}_{dis}$                 & \textbf{0.27}     & \textbf{0.28}     & \emphtable{0.02}                       & \emphtable{31.6}                                                                 & \emphtable{48.3}                                \tabularnewline 
		
		\end{tabular}
\vspace{-0.5em}
		\caption{Loss ablation of Ctrl-SIS on ADE20K. %
  \label{table:loss_ablation}}
		\vspace{-0.5em}
	\end{table}

%% file: tex/conclusion.tex
\section{Conclusion}\label{sec:conclusion}
We propose Ctrl-SIS, which to our knowledge, is the first method for discovering class-specific interpretable GAN controls of SIS models.
This is achieved by optimizing a set of class-specific latent directions via proposed diversity, consistency and disentanglement loss terms, making use of semantic label maps provided as part of the SIS task. The learnt latent directions can locally change the appearance of targeted semantic classes without affecting other classes in the image, and can be combined to sequentially change the image. 
Quantitative and qualitative analysis shows that Ctrl-SIS results in image edits of high quality, that are significantly more diverse than prior methods adapted to SIS.

%% file: appendix.tex
\renewcommand{\thesection}{\Alph{section}}
\renewcommand{\thetable}{\Alph{table}}
\renewcommand{\thefigure}{\Alph{figure}}

\setcounter{figure}{0}   
\setcounter{table}{0}   

\section*{Appendix}

\input{supp_content.tex}

%% file: supp_content.tex
This supplementary material is structured as follows:
\begin{description}
    \item \ref{sec:app_qual} Additional qualitative results.
        \begin{description}
        	\item \ref{sec:app_joint_edit} Visual examples of joint class-specific editing enabled by Ctrl-SIS.
        \item \ref{sec:app_directions} Visual examples of latent directions discovered by Ctrl-SIS.    
        \item \ref{sec:app_related} Qualitative comparison to related work. 
        \end{description}
    \item \ref{sec:app_eval} Details on quantitative evaluation of class-specific GAN controls.
    \item \ref{sec:app_quant} Extended quantitative evaluation. %
    \item \ref{sec:app_limitations} Limitations.%
    
\end{description}

\input{appendix/tex/qual.tex}

\input{appendix/tex/eval.tex}

\input{appendix/tex/quant.tex}

\input{appendix/fig/qual_consistency.tex}

\input{appendix/fig/qual_comparison.tex}

~
\newpage
~
\newpage
~
\newpage

\input{appendix/tex/limitations.tex}

%% file: appendix/tex/qual.tex
\section{Qualitative results}
\label{sec:app_qual}

\subsection{Visual examples of joint class editing}
\label{sec:app_joint_edit}
In Fig.~\ref{fig:joint_1} 
we show examples of images in which two classes are edited jointly with Ctrl-SIS. For example, Fig. \ref{fig:joint_1} shows how bed and curtain, as well as window and wall can be edited separately or jointly. This property is enabled by the use of 3D latent codes, which are spatially aware and can vary image regions independently.

\input{appendix/fig/qual_joint.tex}

\subsection{Visual examples of discovered directions}
\label{sec:app_directions} 
In this section we show latent directions learnt by Ctrl-SIS for the semantic image synthesis model OASIS. We pick OASIS, as it provides the best image quality and diversity (see Table 3 in the main paper). Results on ADE20K and COCO-Stuff are shown in Fig. \ref{fig:app_consistency_A} 
We observe that the directions are consistent across different label maps and change only the image area corresponding to the class of interest. The directions carry different semantics, such as the color of the bus, clouds in the sky, different kinds of house facades, various bed covers, different types of snow, and the lighting of the lamp. %

\subsection{Qualitative comparison to related work} 
\label{sec:app_related}  
Here, we visually compare the diversity between Ctrl-SIS, SeFa and GANSpace. These comparisons are presented in Fig. \ref{fig:app_qual_comparison_1}.
For all methods, the directions are applied to the 3D latent code within the image area corresponding to the selected class. We observe stronger diversity for Ctrl-SIS, which discovers meaningful class-specific directions. For example, in Fig. \ref{fig:app_qual_comparison_1} Ctrl-SIS provides unique views from a window, tree leafage and street surfaces. The stronger diversity is explained by the fact that in contrast to SeFa and GANSpace, Ctrl-SIS is capable of leveraging the label maps that are already available for the task of semantic image synthesis during optimization to learn class-specific directions.

%% file: appendix/fig/qual_joint.tex
\begin{figure*}[h]

    \begin{centering}
    \setlength{\tabcolsep}{0.1em}
    \renewcommand{\arraystretch}{0}
    \par\end{centering} 
    \begin{centering} 
    \hfill{}%
    \begin{tabular}{c@{\hskip 0.1in}c@{\hskip 0.05in}c@{\hskip 0.05in}c@{\hskip 0.05in}c}
     
        \multirow{6}{*}{
            \begin{tabular}{c} 
                \includegraphics[width=0.14\textwidth, height=0.0933\textheight]{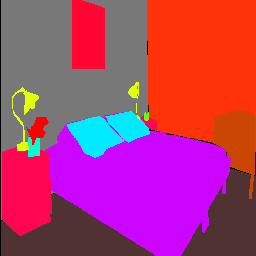}\\
                Label map \\
                \includegraphics[width=0.14\textwidth, height=0.0933\textheight]{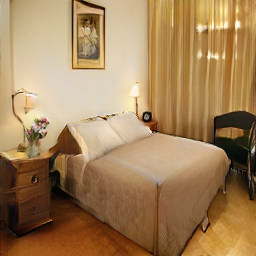} \\
                Generated image \end{tabular}
         } 
        &  %
        \includegraphics[width=0.14\textwidth, height=0.0933\textheight]{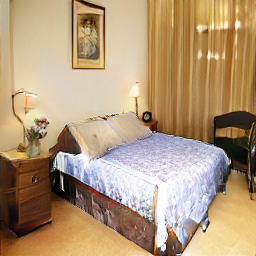} & 
        \includegraphics[width=0.14\textwidth, height=0.0933\textheight]{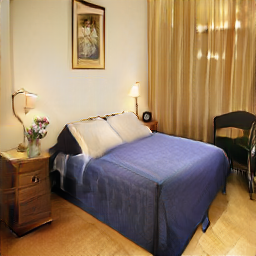} & 
        \includegraphics[width=0.14\textwidth, height=0.0933\textheight]{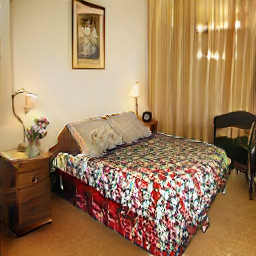} \tabularnewline 
        & 
        \multicolumn{3}{c}{Editing of class \textit{bed}} \tabularnewline 
        & 
        \includegraphics[width=0.14\textwidth, height=0.0933\textheight]{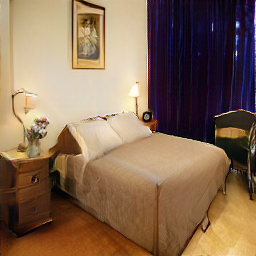} & 
        \includegraphics[width=0.14\textwidth, height=0.0933\textheight]{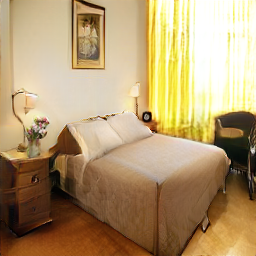} & 
        \includegraphics[width=0.14\textwidth, height=0.0933\textheight]{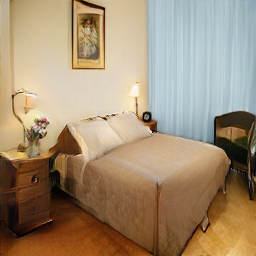} \tabularnewline
        & 
        \multicolumn{3}{c}{Editing of class \textit{curtain}} \tabularnewline
        & 
        \includegraphics[width=0.14\textwidth, height=0.0933\textheight]{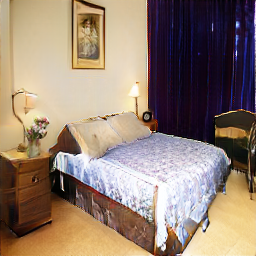} & 
        \includegraphics[width=0.14\textwidth, height=0.0933\textheight]{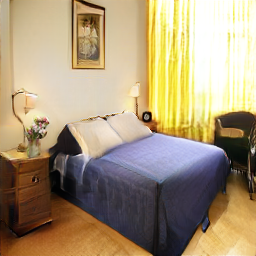} & 
        \includegraphics[width=0.14\textwidth, height=0.0933\textheight]{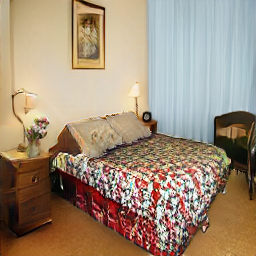} \tabularnewline
        & 
        \multicolumn{3}{c}{Editing of class \textit{bed} and \textit{curtain} jointly}
    \end{tabular}
    \hfill{}
    \par\end{centering} 
    
\end{figure*}

\begin{figure*}[h]
        \begin{centering}
        \setlength{\tabcolsep}{0.1em}
        \renewcommand{\arraystretch}{0}
        \par\end{centering} 
        \begin{centering} 
        \hfill{}%
        \begin{tabular}{c@{\hskip 0.1in}c@{\hskip 0.05in}c@{\hskip 0.05in}c@{\hskip 0.05in}c}
         
            \multirow{6}{*}{
                \begin{tabular}{c} 
                    \includegraphics[width=0.14\textwidth, height=0.0933\textheight]{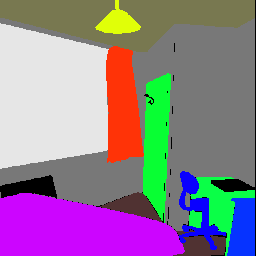}\\
                    Label map \\ 
                    \includegraphics[width=0.14\textwidth, height=0.0933\textheight]{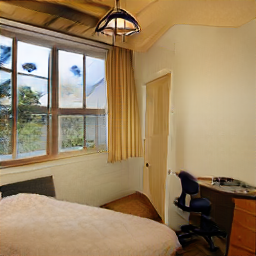} \\
                    Generated image \end{tabular}
             } 
            & 
            \includegraphics[width=0.14\textwidth, height=0.0933\textheight]{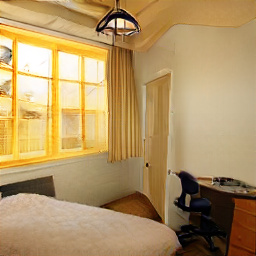} & 
            \includegraphics[width=0.14\textwidth, height=0.0933\textheight]{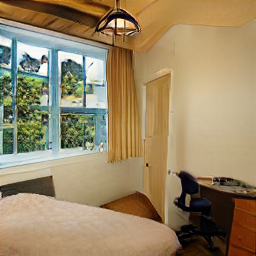} & 
            \includegraphics[width=0.14\textwidth, height=0.0933\textheight]{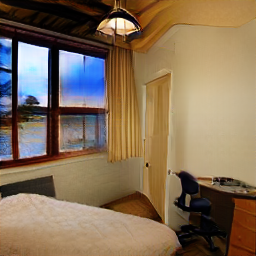} \tabularnewline 
            & 
            \multicolumn{3}{c}{Editing of class \textit{window}} \tabularnewline 
            & 
            \includegraphics[width=0.14\textwidth, height=0.0933\textheight]{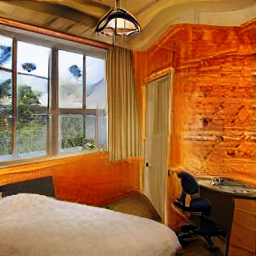} & 
            \includegraphics[width=0.14\textwidth, height=0.0933\textheight]{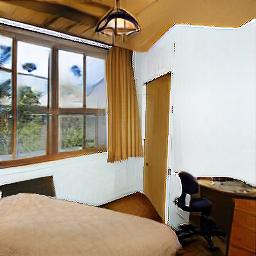} & 
            \includegraphics[width=0.14\textwidth, height=0.0933\textheight]{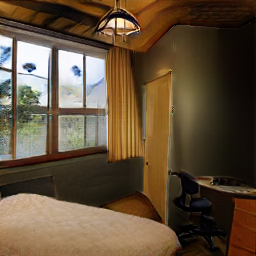} \tabularnewline
            & 
            \multicolumn{3}{c}{Editing of class \textit{wall}} \tabularnewline
            & 
            \includegraphics[width=0.14\textwidth, height=0.0933\textheight]{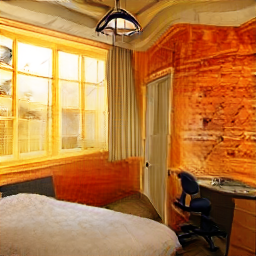} & 
            \includegraphics[width=0.14\textwidth, height=0.0933\textheight]{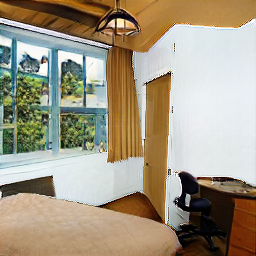} & 
            \includegraphics[width=0.14\textwidth, height=0.0933\textheight]{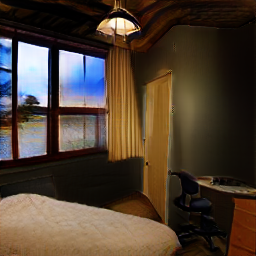} \tabularnewline
            & 
            \multicolumn{3}{c}{Editing of class \textit{window} and \textit{wall} jointly}
        \end{tabular}
        \hfill{}
        \par\end{centering} 
        
        \caption{Joint editing of semantic classes using latent directions learnt by Ctrl-SIS.}
        \label{fig:joint_1} 
\end{figure*}

%% file: appendix/tex/eval.tex
\section{Quantitative evaluation of class-specific GAN controls}
\label{sec:app_eval}
In this section we provide details on our proposed metrics for evaluating GAN controls discovery methods and relate them to prior work~\cite{zhu2020semantically}.
A method for discovering semantically meaningful class-specific directions in the latent space of SIS GANs should exhibit the following three properties: 
First, the found directions should be as unique and different as possible. We assess this property via the \textit{mean control diversity} - mCD.
Second, a latent direction should invoke the same semantic edit independent of the initial latent code, which we assess via the \textit{mean control consistency} - mCC. 
Third, class-specific edits should not affect image areas outside of the target class area. We verify this requirement via the \textit{mean outside class diversity} - mOD. 
The scores are based on computing the LPIPS distance between pairs of images with different edits and the same initial latent code (mCD and mOD), or the same edits but different initial latent codes (mCC). For the global mCD and mCC scores the edits are applied to all classes simultaneously with latent directions that are randomly picked from the set of  discovered class-specific directions.
On the other hand, the local scores mCD$_l$, mCC$_l$ and mOD rely on pairwise distances between images where only one class is edited at a time. 
To compute the pairwise distance between images where only one class is edited, we use the \textit{masked} LPIPS distance. In the following, we explain the masked LPIPS distance and provide the formulations of the local scores mCD$_l$, mCC$_l$ and mOD, as well as the global scores mCD and mCC.

\myparagraph{The masked LPIPS distance.} The default LPIPS distance between two images is based on extracting deep features from both images using a VGG network pretrained on ImageNet classification~\cite{zhang2018perceptual}. The features of all layers are normalized and re-scaled along the channel dimension. The final LPIPS distance is the L2 distance between these features. To compute the \textit{masked} LPIPS distance, we multiply the deep features with a binary mask before computing the L2 distance. We distinguish between LPIPS$^{M_{c}}$ and LPIPS$^{1 - M_{c}}$. The former uses the binary mask $M_c$, which is $1$ where the label map contains class $c$ and $0$ everywhere else. The latter applies the inverted mask $1-M_c$.

\myparagraph{Mean control diversity.}
The mean control diversity is computed for global edits (mCD) and local edits (mCD$_l$).
The mCD$_l$ is computed via:
\begin{equation}
    \mathrm{mCD}_l = \frac{1}{C} \sum_{c=1}^{C}  \mathbb{E}_c \big[ \mathcal{P}_{CD} \big],
\end{equation}
where $C$ is the total number of classes and $\mathcal{P}_{CD}$ denotes the control diversity measured for a label map containing class $c$. 
To compute $\mathcal{P}_{CD}$, a fixed initial latent code is sampled for each label map containing class $c$. Given a label map and its initial latent code, one locally edited image is created for each of the $K$ latent directions specific to class $c$. Next, the average locally masked LPIPS distance is computed between all pairs of the $K$ edited images. This score is averaged over $Z$ initial latent codes, which can be formulated as follows:
\begin{equation}\label{eq:mcd}
    \mathcal{P}_{CD}  =  \frac{1}{ZK}   \sum_z^Z \sum_{ \substack{k_{1,2}=1 \\ k_1\neq k_2}}^K \mathrm{LPIPS}^{M_c}_{z, k_1, k_2}  .
\end{equation}
Here, $\mathrm{LPIPS}^{M_c}_{z, k_1, k_2}$ denotes the LPIPS distance masked with $M_c$ between two images created with the same initial latent code $z$, where class $c$ is edited with latent direction $k_1$ and $k_2$, respectively. 

The mCD for global edits is computed as the average distance between globally edited images on the same label map. For each label map, we create pairs of images with different global edits, changing all classes at once. The class-specific latent directions are randomly chosen for each class. We compute the mean of the default LPIPS distance over all pairs and different initial latent codes. The score is averaged over all label maps in the test set. Higher mCD and mCD$_l$ scores indicate better diversity.

\input{appendix/tables/local_metrics.tex}

\myparagraph{Mean outside class diversity.} 
The spatial disentanglement metric mOD is computed for local edits via
\begin{equation}
    \mathrm{mOD} = \frac{1}{C} \sum_{c=1}^{C}  \mathbb{E}_c \big[ \mathcal{P}_{OD}  \big],
\end{equation}
where $\mathcal{P}_{OD}$ is the outside class diversity measured for a label map containing class $c$.
In contrast to mCD$_l$, the masked LPIPS is computed for the area outside the target class:
\begin{equation}\label{eq:mOD}
    \mathcal{P}_{OD} =  \frac{1}{ZK}   \sum_z^Z \sum_{ \substack{k_{1,2}=1 \\ k_1\neq k_2}}^K \mathrm{LPIPS}^{1-M_c}_{z, k_1, k_2}.
\end{equation}
$\mathrm{LPIPS}^{1-M_c}_{z, k_1, k_2}$ denotes the LPIPS distance masked with $1-M_c$ between two images created with the same initial latent code $z$, where class $c$ is edited locally with the latent direction $k_1$ and $k_2$, respectively. 
A lower mOD indicates better spatial disentanglement.

\myparagraph{Mean control consistency.}
Lastly, to measure the consistency of an edit under different initial latent codes, we compute the mean control consistency for global edits (mCC) and local edits (mCC$_l$). 
The mCC$_l$ is 
\begin{equation}
    \mathrm{mCC}_l = \frac{1}{C} \sum_{c=1}^{C}  \mathbb{E}_c \big[ \mathcal{P}_{CC}  \big],
\end{equation}
where $\mathcal{P}_{CC}$ is the control consistency of a label map containing class $c$.
We compute the pairwise distances between images with different initial latent codes and the same local edit:
\begin{equation}\label{eq:mcc}
    \mathcal{P}_{CC} =  \frac{1}{ZK}   \sum_k^K \sum_{ \substack{z_{1,2}=1 \\ z_1\neq z_2}}^Z  \mathrm{LPIPS}^{M_c}_{k, z_1, z_2} .
\end{equation}
Here, $\mathrm{LPIPS}^{M_c}_{k, z_1, z_2}$ denotes the LPIPS distance masked with $M_c$ between two images created with different initial latent codes $z_1$ and $z_2$, where class $c$ is edited locally with latent direction $k$ for both images.

The global mCC score is computed as the average distance between images with the same global edit but different initial latent codes. For each label map, we create pairs of images with different initial latent codes, but a shared global edit. We compute the mean of the default LPIPS distance over all pairs and across different shared global edits. The score is averaged over all label maps in the test set. Ideally, the mCC and mCC$_l$ are low, indicating high consistency under different initial latent codes.

\input{appendix/tables/table_1.tex} %

\myparagraph{Relation to prior diversity and disentanglement scores.}
The mCD$_l$ and mCC$_l$ are related to the \textit{mean class diversity} (mCSD) and  \textit{mean other class} (mOCD) proposed by~\cite{zhu2020semantically}. These two metrics evaluate diversity and spatial disentanglement for SIS models that allow class-specific manipulations~\cite{zhu2020semantically,schonfeld2020you}. Note that mCSD and mOCD measure the class-specific diversity and disentanglement of a \textit{SIS model}, while our metrics evaluate the class-specific diversity and disentanglement of \textit{a set of discovered latent directions}, allowing us to compare different control discovery methods on the same SIS model. The mCSD measures intra-class diversity as a property of the SIS model itself. In contrast, mCD$_l$ measures the diversity of a set of latent directions, which is a property of the GAN control discovery method. The same relationship holds between mOCD and mOD. 
We next present an extended evaluation using our proposed local metrics mCD$_l$, mCC$_l$ and mOD.

%% file: appendix/tables/local_metrics.tex
\setlength{\tabcolsep}{0.5em}
\renewcommand{\arraystretch}{1}

\begin{table*}[h]

	\centering
	\begin{tabular}{l|ccccc|ccccc}
		
		\multirow{2}{*}{ Method}             & \multicolumn{5}{c|}{ADE20K} & \multicolumn{5}{c}{COCO-Stuff} \tabularnewline
		                                     & \normalsize ${\mathrm{mCD}_l}\uparrow$ & \normalsize ${\mathrm{mCC}_l}\downarrow$                 & \normalsize ${\mathrm{mOD}}\downarrow$ & \normalsize FID$\downarrow$ & \normalsize mIoU$\uparrow$ & \normalsize ${\mathrm{mCD}_l}\uparrow$ & \normalsize ${\mathrm{mCC}_l}\downarrow$ & \normalsize ${\mathrm{mOD}}\downarrow$ & \normalsize FID$\downarrow$ & \normalsize mIoU$\uparrow$\tabularnewline \hline \hline

		Baseline  & -& -& -  & 28.6      & 52.2     & -& -   & - & 17.1  &  42.4
		\tabularnewline
		 \hline
		Random    & 0.04     & 0.17     & \textbf{0.01} & 30.6  & 50.1     & \emphtable{0.02}     &  0.07 &  	\textbf{0.00}	&    17.2   & 44.0
		\tabularnewline

		GANSpace & \emphtable{0.03}    & \textbf{0.15}     & \textbf{0.01} & \textbf{28.3}& \textbf{53.9}       &  \emphtable{0.02}  & \textbf{0.06}   & \textbf{0.00}  &  \textbf{16.7}   &  43.6

		\tabularnewline

		SeFa   & 0.05     & \textbf{0.15}       & \textbf{0.01} & \textbf{28.3}& 53.7     &  \emphtable{0.02}  &  \textbf{0.06} &  \textbf{0.00} & 16.9    & 44.2   
		\tabularnewline

		\textbf{Ctrl-SIS} & \textbf{0.12}       & 0.16       & \textbf{0.01} & 28.8      & 51.6     &\textbf{0.04}      & 0.07		& 0.01   & 17.5    &  \textbf{44.4}
		\tabularnewline
	\end{tabular}
	\caption{Evaluation of OASIS GAN controls on ADE20K and COCO-Stuff on local class-specific edits.\label{table:local}}

\end{table*}

%% file: appendix/tables/table_1.tex
\setlength{\tabcolsep}{0.15em}
\renewcommand{\arraystretch}{1}

\begin{table}[t]

	\centering
	\begin{tabular}{l|ccc|ccc} 
		
		& \multicolumn{3}{c}{\normalsize LPIPS} 
		& \multicolumn{3}{|c}{\normalsize MS-SSIM}\\

		& \footnotesize ${\mathrm{mCD}}\uparrow$ 
		& \footnotesize ${\mathrm{mCC}}\downarrow$   
		& \footnotesize ${\mathrm{mOD}}\downarrow$  
		& \footnotesize ${\mathrm{mCD}}\downarrow$ 
		& \footnotesize ${\mathrm{mCC}}\uparrow$   
		& \footnotesize ${\mathrm{mOD}}\uparrow$  \\
		
		\hline	\hline

		Random             & 0.11     & 0.30     & \textbf{0.01}  & 0.98  & 0.76  & \textbf{1.0}   
		\tabularnewline
		
		GANSpace            & 0.09    & 0.29     & \textbf{0.01} & 0.94  & \textbf{0.78}  & \textbf{1.0}       
		
		\tabularnewline
		
		SeFa                 & 0.12     & \textbf{0.28}       & \textbf{0.01} & 0.92  & \textbf{0.78}  & \textbf{1.0}   
		\tabularnewline
		
		\textbf{Ctrl-SIS}   & \textbf{0.26}       & \textbf{0.28}       & \textbf{0.01} & \textbf{0.74}  & 0.77  & \textbf{1.0}     
		\tabularnewline
	\end{tabular}
	\caption{Evaluation of GAN controls with LPIPS and MS-SSIM using OASIS on ADE20K.\label{table:ctrlsis_msssim}}
\end{table}

 

%% file: appendix/tex/quant.tex
\section{Extended quantitative evaluation}
\label{sec:app_quant}
\subsection{Evaluation on local class-specific edits}
In this section we present an additional comparison between Ctrl-SIS and the related work using OASIS on the ADE20K and COCO-Stuff datasets. For evaluation we employ image quality metrics (FID and mIoU) as well as our proposed diversity (mCD), consistency (mCC) and disentanglement (mOD) scores. In contrast to Table 1 in the main paper, Table \ref{table:local} presents this comparison for \textit{local} edits. While the related work SeFa and GANSpace are designed for global edits, local edits are achieved by adding the learnt global directions to the 3D latent code only in a class-specific image area, as demonstrated in Fig.  \ref{fig:app_qual_comparison_1}. 
As Table \ref{table:local} shows, Ctrl-SIS achieves at least twice the diversity score with respect to SeFa and GANSpace, while the consistency and disentanglement scores stay similar between all methods. The red numbers mark scores which are equal or worse than the ones originating from random directions.
For SeFa and GANSpace, the FID and mIoU are slightly improved compared to unedited images (see Baseline in Table \ref{table:local}), due to generating more "typical" images (see Sec. 4.3 in the main paper). 
In summary, the results from Table \ref{table:local} are in alignment with Table 1 (main paper), suggesting that the editing properties of Ctrl-SIS and related works are similar between local and global edits. 

\subsection{Evaluation with alternative distance measure}
Our proposed scores are based on computing the mean LPIPS distance between pairs of images. 
Here, we also present our metrics computed with the multi-scale structural similarity distance (MS-SSIM)~\cite{wang2003multiscale} as an alternative to LPIPS. 
The main differences between LPIPS and MS-SSIM are as follows. LPIPS computes the L2 distance between image features extracted with a network pre-trained on ImageNet classification. MS-SSIM is not neural network-based and instead computes the similarity between images based on the mean, variance and covariance of two images. A high similarity between two images results in high MS-SSIM but low LPIPS, since LPIPS measures dissimilarity. This means MS-SSIM-based mCD, mCC and mOD scores rise when the LPIPS-based scores fall, and vice versa. 
In Table \ref{table:ctrlsis_msssim}, we compare GAN control discovery methods with our metrics based on LPIPS and MS-SSIM. We note the same trends between the MS-SSIM-based metrics and the LPIPS-based metrics. In particular, Ctrl-SIS also sees a strong increase in diversity under the MS-SSSIM-based mCD metric. The results show that the evaluation metrics are not strictly dependent on the distance measure, and that other ways of estimating image (dis-) similarity may be used.

\subsection{Comparison of GAN control methods across SIS models}
\input{tables/Table3.tex}

In this section, we compare Ctrl-SIS on different SIS models. Table \ref{table:2} shows that Ctrl-SIS exhibits stronger diversity for local and global edits across all tested SIS models. 
The diversity of GANSpace and SeFa is comparable to the diversity measured for random directions (see red numbers in Table \ref{table:2}). 
In other words, the directions that SeFa and GANSpace find differ just as much from each other, as a set of randomly chosen directions. 

A visual comparison of the diversity of Ctrl-SIS, SeFa and GANSpace is shown in figure \ref{fig:app_consistency_A}: In contrast to SeFa and GANSpace, Ctrl-SIS yields latent directions with distinct appearances.  
The directions of GANSpace and SeFa all look very similar. Note that this is comparable to a set of \textit{random} directions. In contrast to regular unconditional or class-conditional GANs, random directions in SIS yield images with low diversity. The low diversity of random directions is a well-known issue for SIS models~\cite{isola2017image,zhu2017toward,schonfeld2020you}.

%% file: tables/Table3.tex
\setlength{\tabcolsep}{0.08em}
\renewcommand{\arraystretch}{0.9}

\begin{table}[t]
	\vspace{-0.5em}
	
	\centering
	\begin{tabular}{l|l|ccc|ccc}
		\multirow{2}{*}{  Model} 
		& \multirow{2}{*}{Method} 
		& \multicolumn{3}{c|}{Global edits} 
		& \multicolumn{3}{c}{Local edits} \tabularnewline
		&                         
		& \footnotesize ${\mathrm{mCD}}$ $\uparrow$       
		& \footnotesize FID $\downarrow$  
		& \footnotesize mIoU $\uparrow$ 
		& \footnotesize ${\mathrm{mCD}_l}$ $\uparrow$ 
		& \footnotesize FID $\downarrow$ 
		& \footnotesize mIoU $\uparrow$\tabularnewline \hline \hline

		\multirow{4}{*}{ OASIS}%
		& Random                  & 0.11                              & 31.3                                                                  & 49.4                       & 0.04                                               & 30.6                                               & 50.1 \tabularnewline
		& GANSpace                & \emphtable{0.09}                              & \textbf{28.1}                                                                     & \textbf{53.3}                       & \emphtable{0.03}                                               & \textbf{28.3}                                                  & \textbf{53.9} \tabularnewline
		& SeFa                    & 0.12                              & \textbf{28.1}                                                                     & 53.2                      & 0.05                                               & \textbf{28.3}                                                  & 53.7 \tabularnewline
		& \textbf{Ctrl-SIS}               & \textbf{0.26}                              & 30.9                                                                     & 48.9                       & \textbf{0.12}                                               & 28.8                                                  & 51.6 \tabularnewline \hline
		\multirow{4}{*}{ SC-GAN}%
		& Random                  & 0.08                              & 34.3                                                                     & 38.1                       & 0.05                                               & \textbf{34.2}                                                  & 38.6			\tabularnewline
		& GANSpace                & 0.11                              & \textbf{34.2}                                                                     & \textbf{38.3}                       & 0.06                                               & 34.3                                                  & 38.8 \tabularnewline
		& SeFa                    & 0.10                              & 34.4                                                                     & 37.8                       & 0.06                                               & 34.4                                                  & \textbf{38.9} \tabularnewline
		& \textbf{Ctrl-SIS}               & \textbf{0.25}                              & 36.4                                                                     & 34.7                       & \textbf{0.18}                                         & \textbf{34.2}                                                  & 38.4  \tabularnewline \hline
		
			\multirow{4}{*}{ SPADE}%
		& Random                  &       0.08                   &   \textbf{34.6}                                                                   &         \textbf{39.4}               &  0.05                                            &        34.6                                           & 39.6		\tabularnewline
		& GANSpace                & 0.12                              & 35.1                                                                    & 39.3                       & 0.08                                               & \textbf{34.6}                                                  & \textbf{39.7}     \tabularnewline
		& SeFa                    & 0.09                              & 34.7                                                                     & \textbf{39.4}                       & 0.06                                               & 34.8                                                  & \textbf{39.7}     \tabularnewline
		& \textbf{Ctrl-SIS}               & \textbf{0.14}                              & 35.4                                                                    & 38.6                       & \textbf{0.09}                                              & \textbf{34.6}                                                  & 39.4     \tabularnewline
	\end{tabular}
	\vspace{-0.5em}
	\caption{Comparison of GAN control methods across SIS models on ADE20K.\label{table:2}}
	\vspace{-0.5em}
\end{table}

%% file: appendix/fig/qual_consistency.tex
\begin{figure*}[h]
	\begin{centering}
		\setlength{\tabcolsep}{0.1em}
		\renewcommand{\arraystretch}{0}
		\par\end{centering}
	\begin{centering}
	
		\hfill{}
		
		\begin{tabular}{c@{\hskip 0.042in} c@{\hskip 0.042in} c@{\hskip 0.042in} c@{\hskip 0.042in} c@{\hskip 0.042in} c@{\hskip 0.042in} c@{\hskip 0.008in} c}

			\includegraphics[width=0.14\textwidth, height=0.0933\textheight]{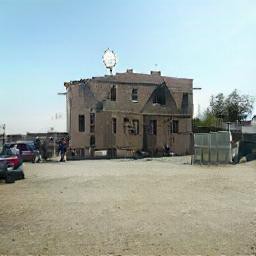} & 
			\includegraphics[width=0.14\textwidth, height=0.0933\textheight]{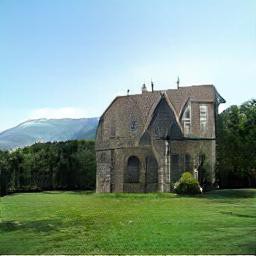} & 
			\includegraphics[width=0.14\textwidth, height=0.0933\textheight]{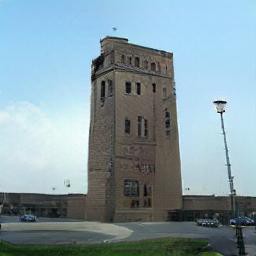} & 
			\includegraphics[width=0.14\textwidth, height=0.0933\textheight]{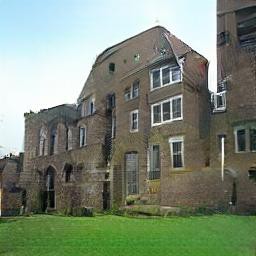} & 
			\includegraphics[width=0.14\textwidth, height=0.0933\textheight]{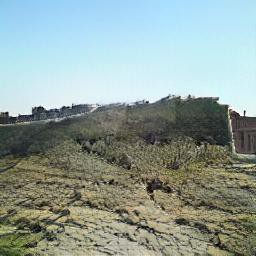} & 
			\includegraphics[width=0.14\textwidth, height=0.0933\textheight]{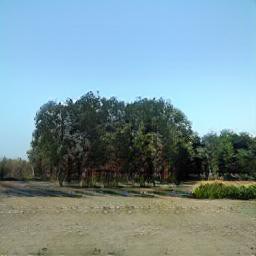} & 
			\\ 
			\includegraphics[width=0.14\textwidth, height=0.0933\textheight]{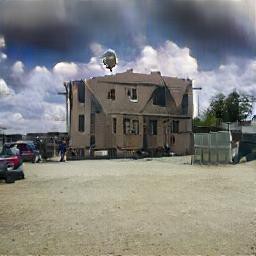} & 
			\includegraphics[width=0.14\textwidth, height=0.0933\textheight]{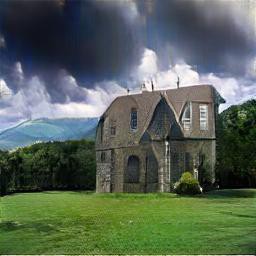} & 
			\includegraphics[width=0.14\textwidth, height=0.0933\textheight]{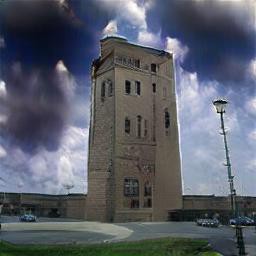} & 
			\includegraphics[width=0.14\textwidth, height=0.0933\textheight]{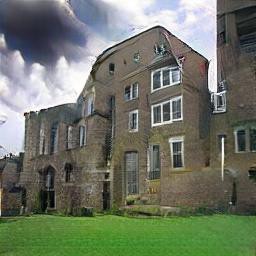} & 
			\includegraphics[width=0.14\textwidth, height=0.0933\textheight]{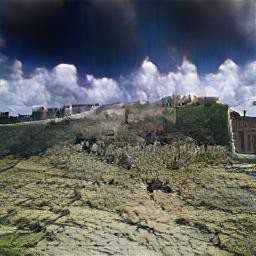} & 
			\includegraphics[width=0.14\textwidth, height=0.0933\textheight]{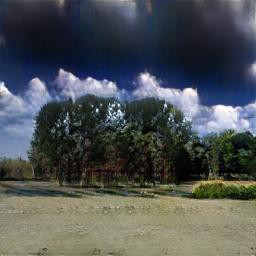} & 
			\\ 
			
				\multicolumn{6}{c}{ sky} \vspace{0.8em}
			\\ 

    		\includegraphics[width=0.14\textwidth, height=0.0933\textheight]{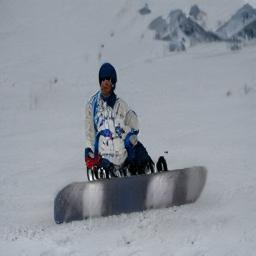} & 
    		\includegraphics[width=0.14\textwidth, height=0.0933\textheight]{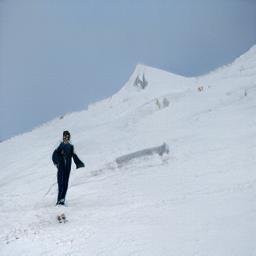} & 
    		\includegraphics[width=0.14\textwidth, height=0.0933\textheight]{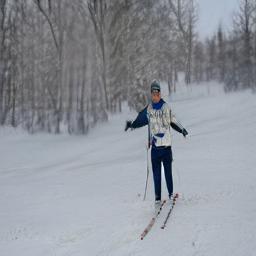} & 
    		\includegraphics[width=0.14\textwidth, height=0.0933\textheight]{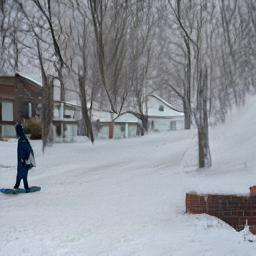} & 
    		\includegraphics[width=0.14\textwidth, height=0.0933\textheight]{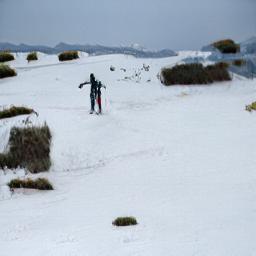} & 
    		\includegraphics[width=0.14\textwidth, height=0.0933\textheight]{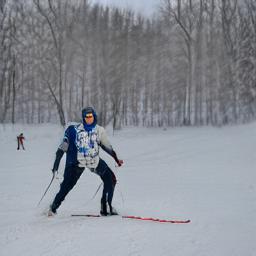} & 
    		\\ 
    		\includegraphics[width=0.14\textwidth, height=0.0933\textheight]{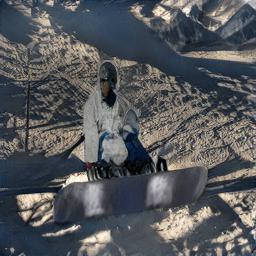} & 
    		\includegraphics[width=0.14\textwidth, height=0.0933\textheight]{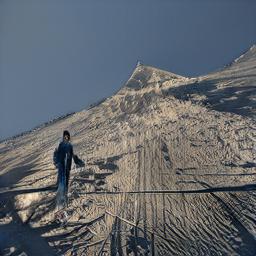} & 
    		\includegraphics[width=0.14\textwidth, height=0.0933\textheight]{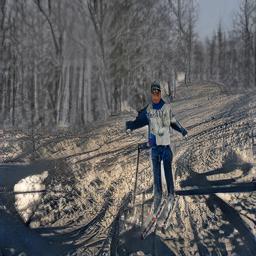} & 
    		\includegraphics[width=0.14\textwidth, height=0.0933\textheight]{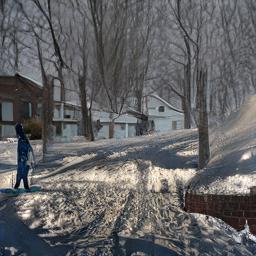} & 
    		\includegraphics[width=0.14\textwidth, height=0.0933\textheight]{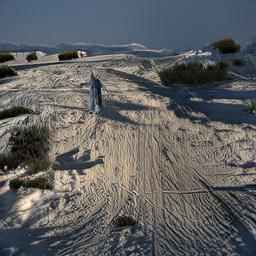} & 
    		\includegraphics[width=0.14\textwidth, height=0.0933\textheight]{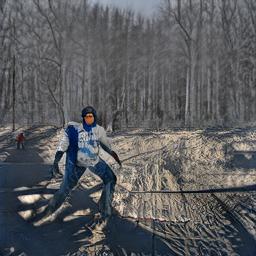} & 
    		\\
    		
    		\multicolumn{6}{c}{snow} \vspace{0.8em}
    		\\

			\includegraphics[width=0.14\textwidth, height=0.0933\textheight]{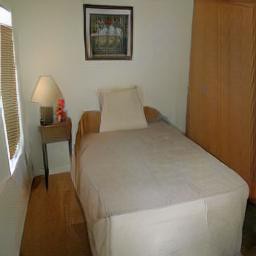} & %
			\includegraphics[width=0.14\textwidth, height=0.0933\textheight]{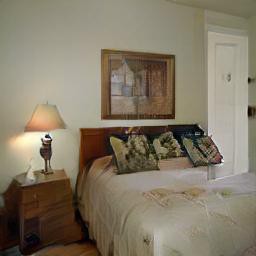} & 
			\includegraphics[width=0.14\textwidth, height=0.0933\textheight]{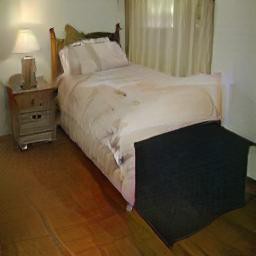} & 
			\includegraphics[width=0.14\textwidth, height=0.0933\textheight]{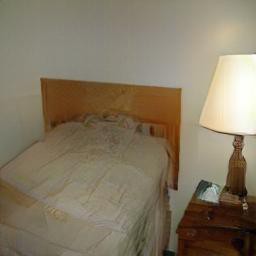} & %
			\includegraphics[width=0.14\textwidth, height=0.0933\textheight]{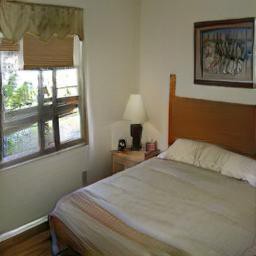} & %
			\includegraphics[width=0.14\textwidth, height=0.0933\textheight]{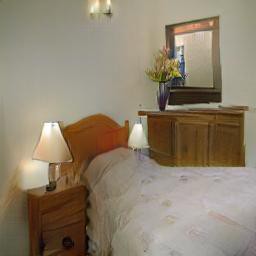} & 
			\\
			\includegraphics[width=0.14\textwidth, height=0.0933\textheight]{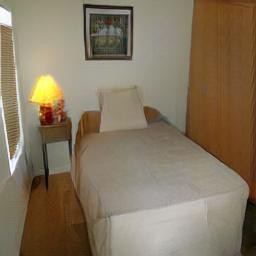} & %
			\includegraphics[width=0.14\textwidth, height=0.0933\textheight]{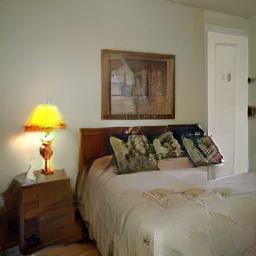} & 
			\includegraphics[width=0.14\textwidth, height=0.0933\textheight]{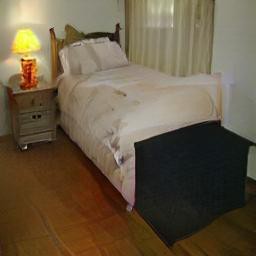} & 
			\includegraphics[width=0.14\textwidth, height=0.0933\textheight]{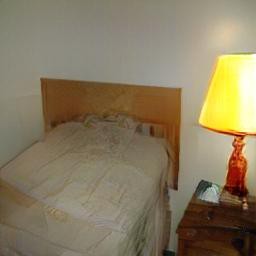} & %
			\includegraphics[width=0.14\textwidth, height=0.0933\textheight]{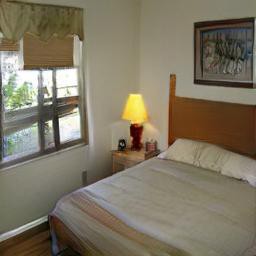} & %
			\includegraphics[width=0.14\textwidth, height=0.0933\textheight]{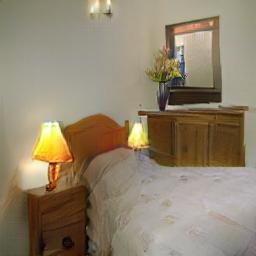} & 
			\\ 
			\multicolumn{6}{c}{lamp} \vspace{0.8em}
			\\ 
		
			\includegraphics[width=0.14\textwidth, height=0.0933\textheight]{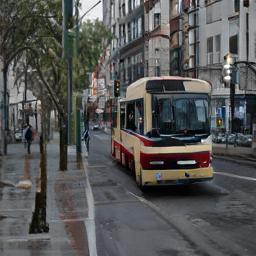} & 
			\includegraphics[width=0.14\textwidth, height=0.0933\textheight]{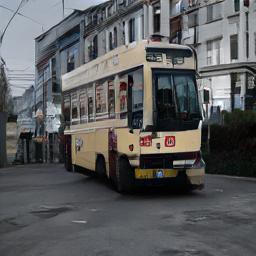} & 
			\includegraphics[width=0.14\textwidth, height=0.0933\textheight]{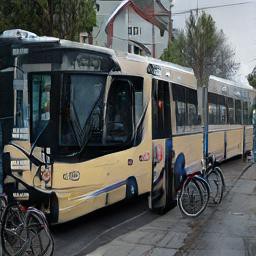} & 
			\includegraphics[width=0.14\textwidth, height=0.0933\textheight]{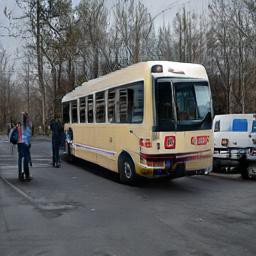} & 
			\includegraphics[width=0.14\textwidth, height=0.0933\textheight]{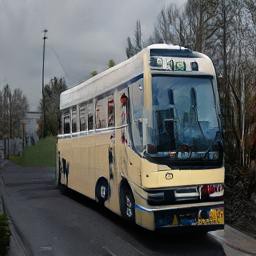} & 
			\includegraphics[width=0.14\textwidth, height=0.0933\textheight]{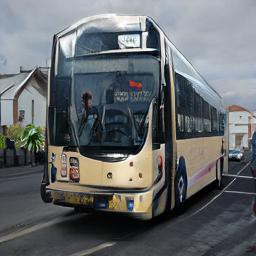} &
			\\
			\includegraphics[width=0.14\textwidth, height=0.0933\textheight]{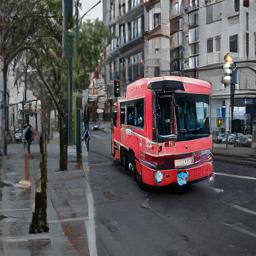} & 
			\includegraphics[width=0.14\textwidth, height=0.0933\textheight]{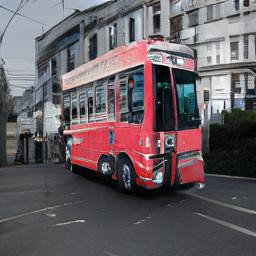} & 
			\includegraphics[width=0.14\textwidth, height=0.0933\textheight]{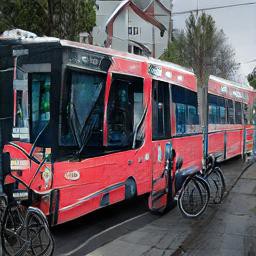} & 
			\includegraphics[width=0.14\textwidth, height=0.0933\textheight]{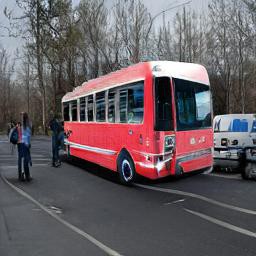} & 
			\includegraphics[width=0.14\textwidth, height=0.0933\textheight]{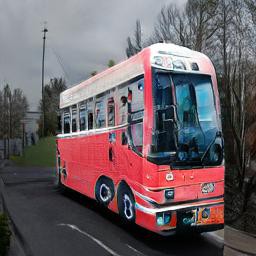} & 
			\includegraphics[width=0.14\textwidth, height=0.0933\textheight]{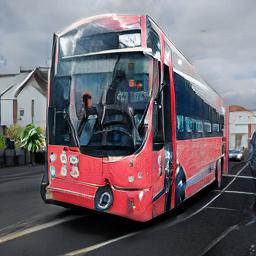} & 
			
			\\ 
			\multicolumn{6}{c}{bus} \vspace{0.8em}
			\\

		\end{tabular}
		
		\par\end{centering}

	\caption{Latent directions learnt by Ctrl-SIS on ADE20K and COCO-Stuff.} 
	\label{fig:app_consistency_A}
\end{figure*}

%% file: appendix/fig/qual_comparison.tex
\begin{figure*}[h]
	\begin{centering}
		\setlength{\tabcolsep}{0.1em}
		\renewcommand{\arraystretch}{0}
		\par\end{centering}
	\begin{centering}
		\vspace{-1em}
		\begin{tabular}{@{}c@{\hskip 0.1in} c@{\hskip 0.04in} c@{\hskip 0.04in} c@{\hskip 0.04in} c@{\hskip 0.04in} c@{\hskip 0.04in} c@{}c@{}}

\multirow{1}{*}{ \rotatebox{90}{ \small GANSpace \hspace{3.3em} SeFa \hspace{3.7em} Ctrl-SIS \hspace{-4.2em}}}   &
		\includegraphics[width=0.14\textwidth, height=0.0933\textheight]{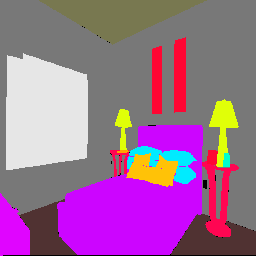} & 
		\includegraphics[width=0.14\textwidth, height=0.0933\textheight]{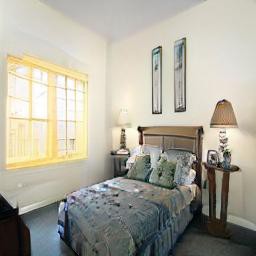} & 
		\includegraphics[width=0.14\textwidth, height=0.0933\textheight]{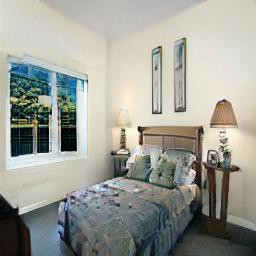} & 
		\includegraphics[width=0.14\textwidth, height=0.0933\textheight]{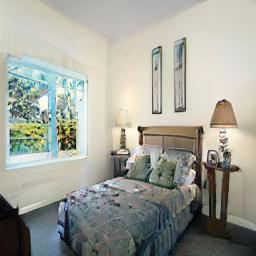} & 
		\includegraphics[width=0.14\textwidth, height=0.0933\textheight]{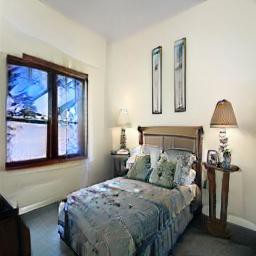} & 
		\includegraphics[width=0.14\textwidth, height=0.0933\textheight]{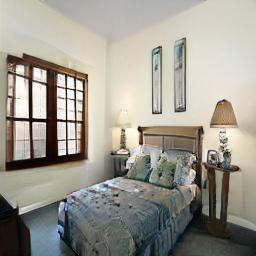} %
		\\
		&
		\includegraphics[width=0.14\textwidth, height=0.0933\textheight]{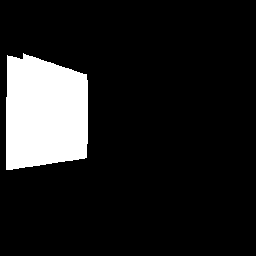} & 
		\includegraphics[width=0.14\textwidth, height=0.0933\textheight]{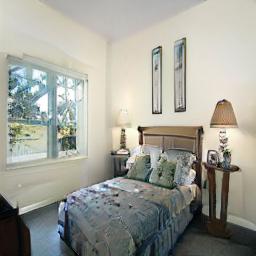} & 
		\includegraphics[width=0.14\textwidth, height=0.0933\textheight]{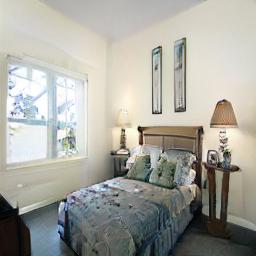} & 
		\includegraphics[width=0.14\textwidth, height=0.0933\textheight]{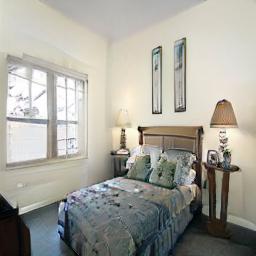} & 
		\includegraphics[width=0.14\textwidth, height=0.0933\textheight]{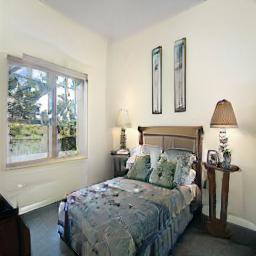} & 
		\includegraphics[width=0.14\textwidth, height=0.0933\textheight]{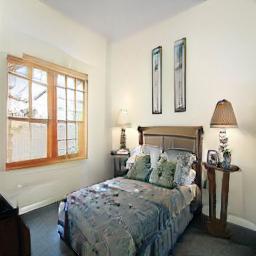} 
		\\
		&
		\includegraphics[width=0.14\textwidth, height=0.0933\textheight]{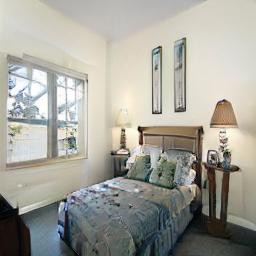} & 
		\includegraphics[width=0.14\textwidth, height=0.0933\textheight]{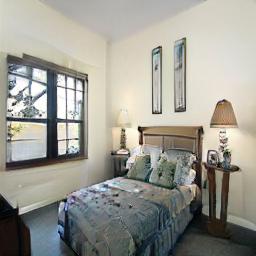} & 
		\includegraphics[width=0.14\textwidth, height=0.0933\textheight]{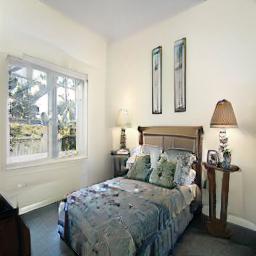} & 
		\includegraphics[width=0.14\textwidth, height=0.0933\textheight]{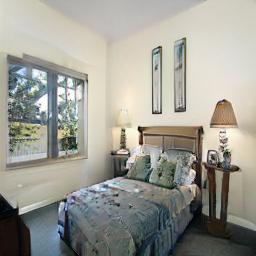} & 
		\includegraphics[width=0.14\textwidth, height=0.0933\textheight]{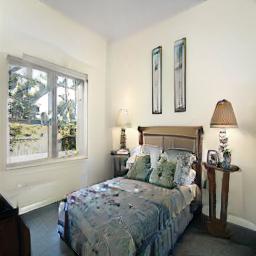} & 
		\includegraphics[width=0.14\textwidth, height=0.0933\textheight]{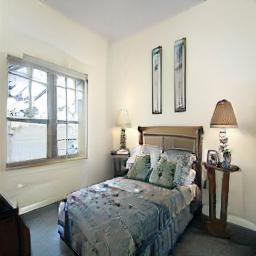} 
		\\
		& \multicolumn{6}{c}{window}  \vspace{0.5em} 
		\\

    			   \multirow{1}{*}{ \rotatebox{90}{ \small GANSpace \hspace{3.3em} SeFa \hspace{3.7em} Ctrl-SIS \hspace{-4.2em}}}   &
    			    \includegraphics[width=0.14\textwidth, height=0.0933\textheight]{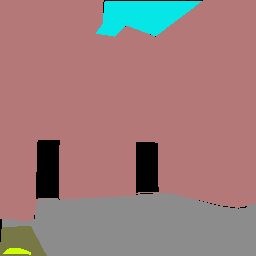} & 
    				\includegraphics[width=0.14\textwidth, height=0.0933\textheight]{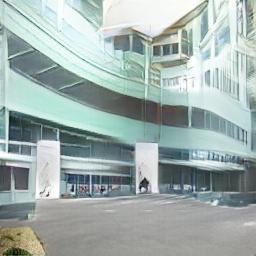} & 
    				\includegraphics[width=0.14\textwidth, height=0.0933\textheight]{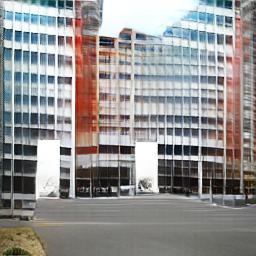} & 
    				\includegraphics[width=0.14\textwidth, height=0.0933\textheight]{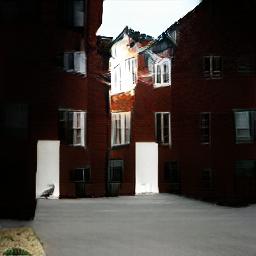} & 
    				\includegraphics[width=0.14\textwidth, height=0.0933\textheight]{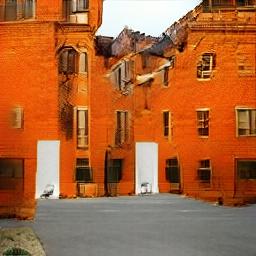} & 
    				\includegraphics[width=0.14\textwidth, height=0.0933\textheight]{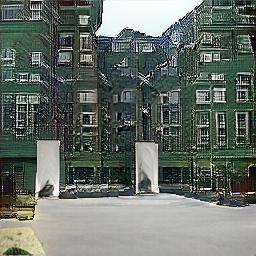} %
    				\\
    				&
    				\includegraphics[width=0.14\textwidth, height=0.0933\textheight]{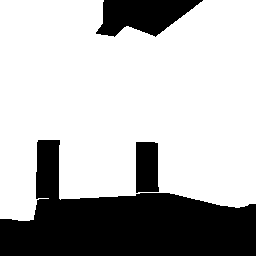} & 
    				\includegraphics[width=0.14\textwidth, height=0.0933\textheight]{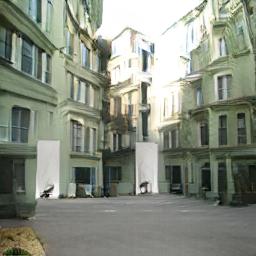} & 
    				\includegraphics[width=0.14\textwidth, height=0.0933\textheight]{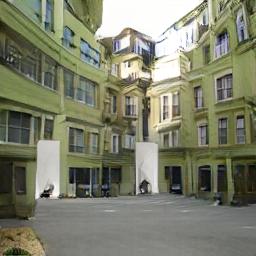} & 
    				\includegraphics[width=0.14\textwidth, height=0.0933\textheight]{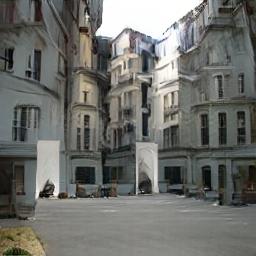} & 
    				\includegraphics[width=0.14\textwidth, height=0.0933\textheight]{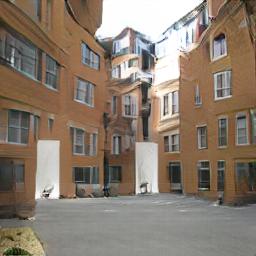} & 
    				\includegraphics[width=0.14\textwidth, height=0.0933\textheight]{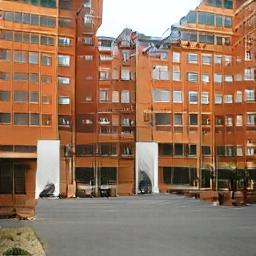} 
    				\\
    				&
    				\includegraphics[width=0.14\textwidth, height=0.0933\textheight]{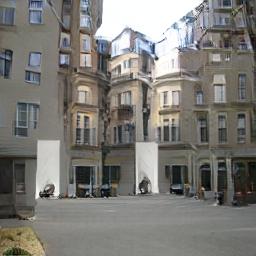} & 
    				\includegraphics[width=0.14\textwidth, height=0.0933\textheight]{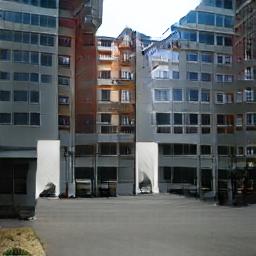} & 
    				\includegraphics[width=0.14\textwidth, height=0.0933\textheight]{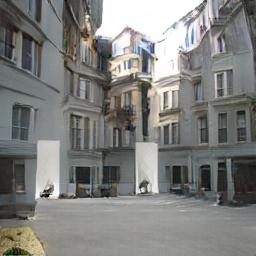} & 
    				\includegraphics[width=0.14\textwidth, height=0.0933\textheight]{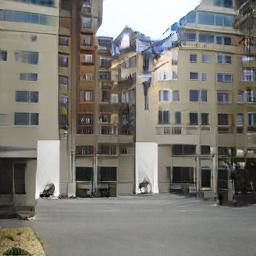} & 
    				\includegraphics[width=0.14\textwidth, height=0.0933\textheight]{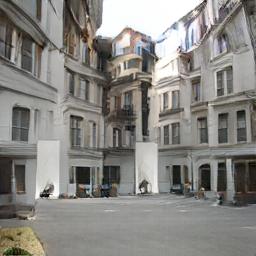} & 
    				\includegraphics[width=0.14\textwidth, height=0.0933\textheight]{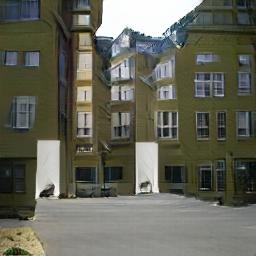} 
    				\\
    				& \multicolumn{6}{c}{house} \vspace{0.5em} 
    				\\
    				
    		\multirow{1}{*}{ \rotatebox{90}{ \small GANSpace \hspace{3.3em} SeFa \hspace{3.7em} Ctrl-SIS \hspace{-4.2em}}}   &
    			\includegraphics[width=0.14\textwidth, height=0.0933\textheight]{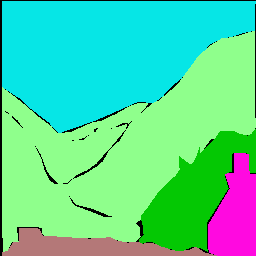} & 
    			\includegraphics[width=0.14\textwidth, height=0.0933\textheight]{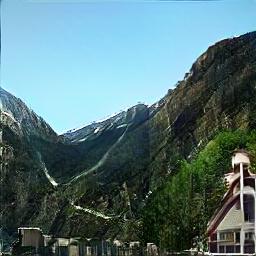} & 
    			\includegraphics[width=0.14\textwidth, height=0.0933\textheight]{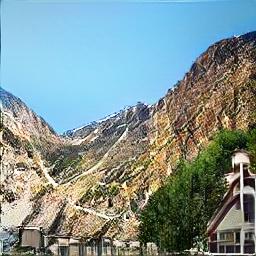} & 
    			\includegraphics[width=0.14\textwidth, height=0.0933\textheight]{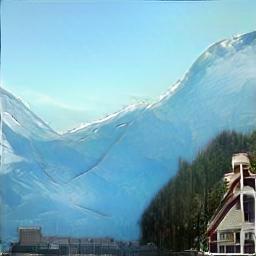} & 
    			\includegraphics[width=0.14\textwidth, height=0.0933\textheight]{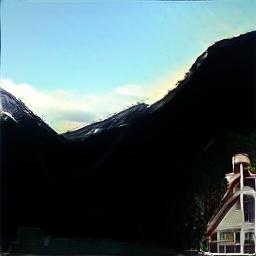} & 
    			\includegraphics[width=0.14\textwidth, height=0.0933\textheight]{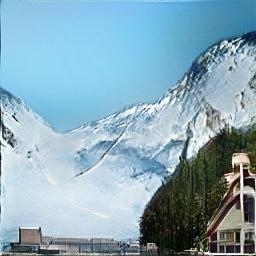} %
    			\\
    			&
    			\includegraphics[width=0.14\textwidth, height=0.0933\textheight]{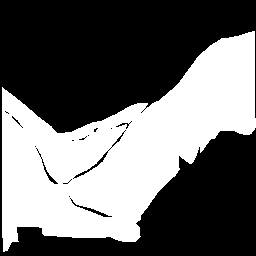} & 
    			\includegraphics[width=0.14\textwidth, height=0.0933\textheight]{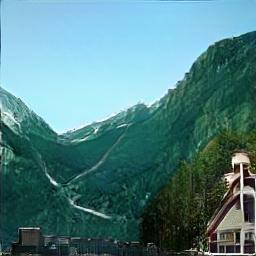} & 
    			\includegraphics[width=0.14\textwidth, height=0.0933\textheight]{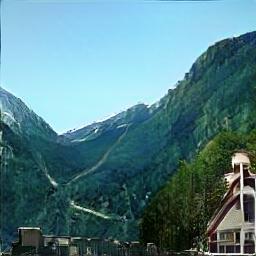} & 
    			\includegraphics[width=0.14\textwidth, height=0.0933\textheight]{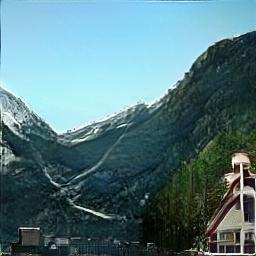} & 
    			\includegraphics[width=0.14\textwidth, height=0.0933\textheight]{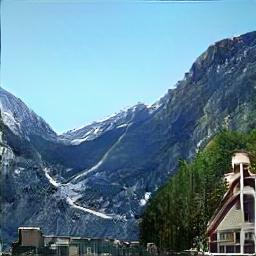} & 
    			\includegraphics[width=0.14\textwidth, height=0.0933\textheight]{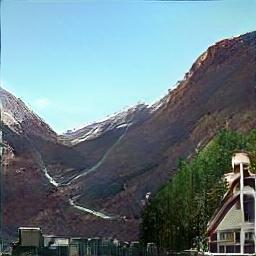} 
    			\\
    			&
    			\includegraphics[width=0.14\textwidth, height=0.0933\textheight]{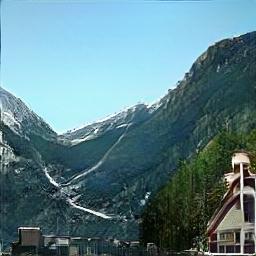} & 
    			\includegraphics[width=0.14\textwidth, height=0.0933\textheight]{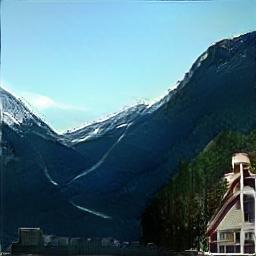} & 
    			\includegraphics[width=0.14\textwidth, height=0.0933\textheight]{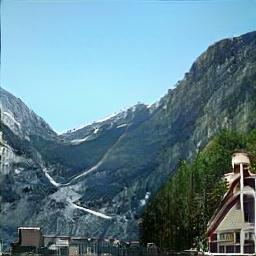} & 
    			\includegraphics[width=0.14\textwidth, height=0.0933\textheight]{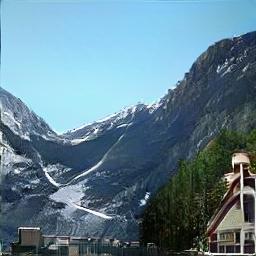} & 
    			\includegraphics[width=0.14\textwidth, height=0.0933\textheight]{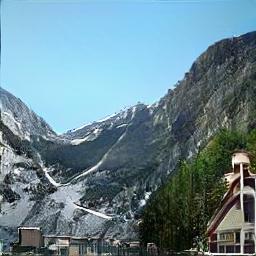} & 
    			\includegraphics[width=0.14\textwidth, height=0.0933\textheight]{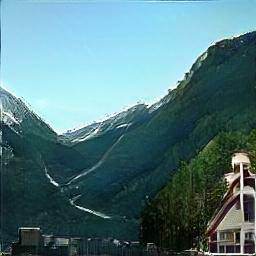} 
    			\\
    			& \multicolumn{6}{c}{mountain} \vspace{0.5em} 
    		       \\

		\end{tabular}

		\par\end{centering}
	\vspace{-0.5em}
	\caption{Qualitative comparison of Ctrl-SIS against SeFa and GANSpace.}
	\label{fig:app_qual_comparison_1}
\end{figure*}

%% file: appendix/tex/limitations.tex
\section{Limitations}
\label{sec:app_limitations}

There are two main limitations to what a method for class-specific latent direction discovery can do. 
First, class-specific directions do not encode shape-based semantics. For example, there cannot be a class-specific direction encoding a "smile" in a face dataset if the shape of the mouth is already hard-coded by the label map.
Second, the diversity of Ctrl-SIS is limited by the diversity of the SIS model to which it is applied. Notably, the diversity of SIS models is far lower than the diversity of regular unconditional or class-conditional GANs. While a standard unconditional GAN produces seemingly infinitely many different images, the diversity of SIS models like OASIS~\cite{schonfeld2020you} was limited to a manageable number of distinct appearances, based on our experience. The problem of diversity in SIS models is a well-known problem ~\cite{isola2017image,zhu2017toward,schonfeld2020you}. Consequently, more diverse SIS models will lead to more diverse sets of discovered latent directions.